\documentclass[sigconf]{acmart}

\copyrightyear{2024}
\acmYear{2024}
\setcopyright{acmlicensed}
\acmConference[KDD '24]{Proceedings of the 30th ACM SIGKDD Conference on Knowledge Discovery and Data Mining}{August 25--29, 2024}{Barcelona, Spain}
\acmBooktitle{Proceedings of the 30th ACM SIGKDD Conference on Knowledge Discovery and Data Mining (KDD '24), August 25--29, 2024, Barcelona, Spain}
\acmISBN{979-8-4007-0490-1/24/08}
\acmDOI{10.1145/3637528.3671913}

\AtBeginDocument{%
  \providecommand\BibTeX{{%
    \normalfont B\kern-0.5em{\scshape i\kern-0.25em b}\kern-0.8em\TeX}}}

\usepackage{tcolorbox}
\usepackage{enumitem}
\setlist[itemize]{leftmargin=*}
\usepackage{booktabs} 
\usepackage{siunitx} 
\usepackage{multirow}
\usepackage[ruled,linesnumbered]{algorithm2e} 

\begin{document}


\title{All in One and One for All: A Simple yet Effective Method towards Cross-domain Graph Pretraining}



\author{Haihong Zhao}
\authornote{Equal Contribution.}
\affiliation{
  \institution{Data Science and Analytics Thrust, The Hong Kong University of Science and Technology (Guangzhou)}
  \streetaddress{}
  \city{Guangzhou}
  \country{China}
  }
\email{hzhaobf@connect.ust.hk}

\author{Aochuan Chen}
\authornotemark[1]
\affiliation{
  \institution{Data Science and Analytics Thrust, The Hong Kong University of Science and Technology (Guangzhou)}
  \streetaddress{}
  \city{Guangzhou}
  \country{China}
  }
\email{achen149@connect.hkust-gz.edu.cn}


\author{Xiangguo Sun}
\authornotemark[1]
\authornote{Corresponding author.}
\affiliation{
  \institution{Department of Systems Engineering and Engineering Management, and Shun Hing Institute of Advanced Engineering, The Chinese University of Hong Kong}
  \streetaddress{}
  \city{Hong Kong SAR}
  \country{China}
  }
\email{xgsun@se.cuhk.edu.hk}

\author{Hong Cheng}
\affiliation{
  \institution{Department of Systems Engineering and Engineering Management, and Shun Hing Institute of Advanced Engineering, The Chinese University of Hong Kong}
  \streetaddress{}
  \city{Hong Kong SAR}
  \country{China}
  }
\email{hcheng@se.cuhk.edu.hk}

\author{Jia Li}
\authornotemark[2]
\affiliation{
  \institution{Data Science and Analytics Thrust, The Hong Kong University of Science and Technology (Guangzhou)}
  \streetaddress{}
  \city{Guangzhou}
  \country{China}
  }
\email{jialee@ust.hk}

\renewcommand{\shortauthors}{Haihong Zhao, Aochuan Chen, Xiangguo Sun, Hong Cheng \& Jia Li}
\vspace{-1em}







\begin{abstract}
    Large Language Models (LLMs) have revolutionized the fields of computer vision (CV) and natural language processing (NLP). 
    One of the most notable advancements of LLMs is that a single model is trained on vast and diverse datasets spanning multiple domains -- a paradigm we term `All in One'. This methodology empowers LLMs with super generalization capabilities, facilitating an encompassing comprehension of varied data distributions. Leveraging these capabilities, a single LLM demonstrates remarkable versatility across a variety of domains -- a paradigm we term `One for All'.
    However, applying this idea to the graph field remains a formidable challenge, with cross-domain pretraining often resulting in negative transfer. This issue is particularly important in few-shot learning scenarios, where the paucity of training data necessitates the incorporation of external knowledge sources.
    In response to this challenge, we propose a novel approach called Graph COordinators for PrEtraining (GCOPE), that harnesses the underlying commonalities across diverse graph datasets to enhance few-shot learning. Our novel methodology involves a unification framework that amalgamates disparate graph datasets during the pretraining phase to distill and transfer meaningful knowledge to target tasks. 
    Extensive experiments across multiple graph datasets demonstrate the superior efficacy of our approach. By successfully leveraging the synergistic potential of multiple graph datasets for pretraining, our work stands as a pioneering contribution to the realm of graph foundational model. 
    Code available at \textbf{\url{https://github.com/cshhzhao/GCOPE}}.
\end{abstract}






\maketitle
\section{Introduction}\label{sec:intro}

Recently, artificial general intelligence (AGI) has achieved remarkable advancement in the realms of computer vision (CV) \cite{chen2023understanding, zhang2023selectivity, chen2020simple,ridnik2021imagenet} and natural language processing (NLP) \cite{touvron2023llama, du2021all}. One of the key innovations for these models is that they pre-train one foundation model through various contexts, making the model absorb and synthesize knowledge across diverse domains (a.k.a ``all in one'') to deliver robust, context-aware responses. Their ability to generalize and apply learned knowledge to a wide spectrum of downstream domains (a.k.a ``one for all'') is a testament to the success of their pre-training strategies, which effectively capture and utilize the nuances across different domains.

Compared with NLP and CV areas, graph data is non-linear and more general. For example, a sentence can be seen as a graph path and an image can be seen as a grid graph. These observations indicate a broader and more ambitious goal of a more generalized and versatile artificial intelligence by graphs. However, pre-training on graphs is still grappling with the integration of multi-domain knowledge and cross-domain generalization. Although existing graph transfer learning methods  \cite{zhuang2020comprehensive,zhao2023kegnn,sun2023all,sun2023graph,li2023survey, gprompt23} demonstrate proficiency within the same domain, their successes, often confined to transferring tasks within a similar graph dataset, exhibit inherent limitations when attempting to transcend the graph domain boundary. 
These disparities make it difficult for graph models to replicate the success of their NLP and CV counterparts, yet the pursuit of moving beyond existing domain-specific task transferring and towards a more versatile approach is not just an extension of existing methodologies but a stride towards bridging the gap between domain-specific proficiency and the broader vision of general artificial intelligence.


However, achieving this vision is very intractable. \textbf{The first challenge} lies in the diversity of structural patterns, which is particularly observed between homophilic graphs (a pair of nodes are intended to be similar if they are connected) and heterophilic graphs (connected nodes depart from each other). 
Graphs inherently exhibit diverse topologies and features, making it challenging to identify and leverage common patterns across different domains. The good thing is that this diversity enriches the external knowledge and contains huge potential for more general transferability to various downstream applications. However, it also complicates the task of identifying and leveraging commonalities across domains, as models must balance the fine line between capturing the essence of individual graphs and maintaining the generalization across different datasets.

\textbf{The second challenge} is the complexity inherent in aligning semantic spaces (features) across different graph datasets. Unlike domains with a structured, common framework (like images or texts), graph data is diverse and abstract. Features in one graph might have no direct counterpart in another, making it incredibly challenging to align these features in a meaningful way. This alignment is not just a matter of mapping features but of understanding and reconciling the underlying semantics and contexts they represent. The absence of a universal framework to guide this alignment makes it a daunting task, one that requires a sophisticated approach to bridge the gap between disparate graph datasets effectively. Some recent works \cite{liu2023one} are trying to align different graph semantic spaces by generating a textual description of these graphs and then using LLMs to learn a unified representation, but they can only deal with text-attributed graphs or graphs with specific feature semantics. In a broader range, many graphs have only latent feature vectors and we actually do not know how exactly each dimension means, let alone generate a textual description.

Inspired by recent achievements of graph prompt \cite{sun2023all}, which has been carefully demonstrated as a powerful approach to manipulate various graph data, we hope to introduce the graph prompting technique to conduct more advanced operations on different graph datasets, to bridge the gap left by cross-domain graph pre-training and transferring. Specifically, we introduce the concept of ``coordinators'', which are some virtual nodes that function as dynamic bridges between disparate graph datasets. These coordinators share the same theory foundation of graph prompts that have the powerful capability of data manipulation. The theory foundation ensures that these coordinators can learn an appropriate latent data manipulation strategy to harmonize different datasets, promote a unified representation, and adapt to the specificities of each graph, thus ensuring that the model doesn't just see a collection of isolated datasets but a harmonious, interconnected network. This innovation allows the model to navigate the diversity of graph structures adeptly, recognizing and leveraging underlying commonalities and differences in a balanced and informed manner. Beyond the coordinators, we also design a complete pretraining framework and provide two transferring components, which can ensure that the knowledge transferred is not just relevant but also contextually enriched, turning potential negative transfer into a positive, performance-enhancing phenomenon. In summary, the main contributions are as follows:
\begin{itemize}
    \item We propose a very simple yet very effective framework, named ``GCOPE'', as a novel mechanism to unify diverse graph structures, enabling the seamless integration and a unified graph pre-training framework, making the pre-trained graph model able to preserve multi-domain knowledge across different graph datasets. 
    \item Inspired by the powerful data manipulation capability of graph prompts in the downstream stage, we propose similar  ``coordinators'' in the pre-training stage so that we can learn a latent data alignment strategy and ensure the relevance and efficacy of knowledge transfer, especially in few-shot learning scenarios. 
  \item Our pre-training framework is compatible with various graph pre-training methods and the transferring stage can also leverage both fine-tuning and prompting techniques.
  \item We conduct comprehensive experimental evaluation, the results from which demonstrate the superior performance of our approach across multiple graph datasets.
\end{itemize}



\section{Motivation}
\label{sec:motivation}

\begin{figure}[htb]
    \vspace{-1.15em}
    \includegraphics[width=0.46\textwidth]{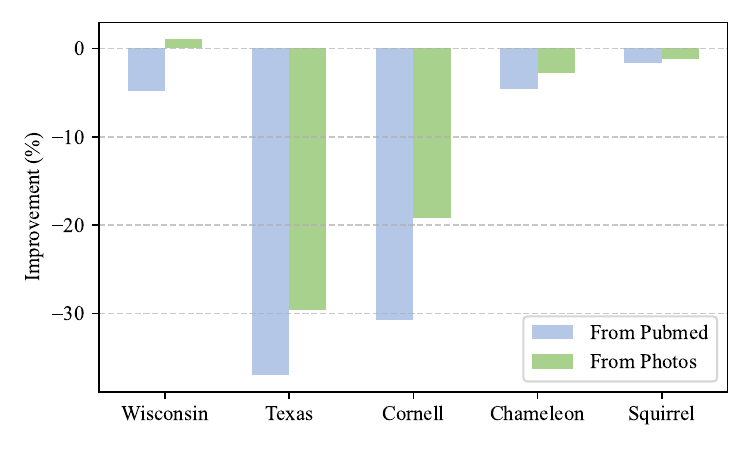}
    \vspace{-1.5em}
    \caption{Negative transfer phenomenon in the single-source cross-domain transfer setting. Sources (Pubmed and Photos) are two homophilic datasets. Targets (Wisconsin, Texas, Cornell, Chameleon, and Squirrel) are five heterophilic graphs.}
    \label{fig: negative transfer analysis}    
    \vspace{-0.65em}
\end{figure}

\begin{figure*}[htb]
    \centering
    \includegraphics[width=\textwidth]{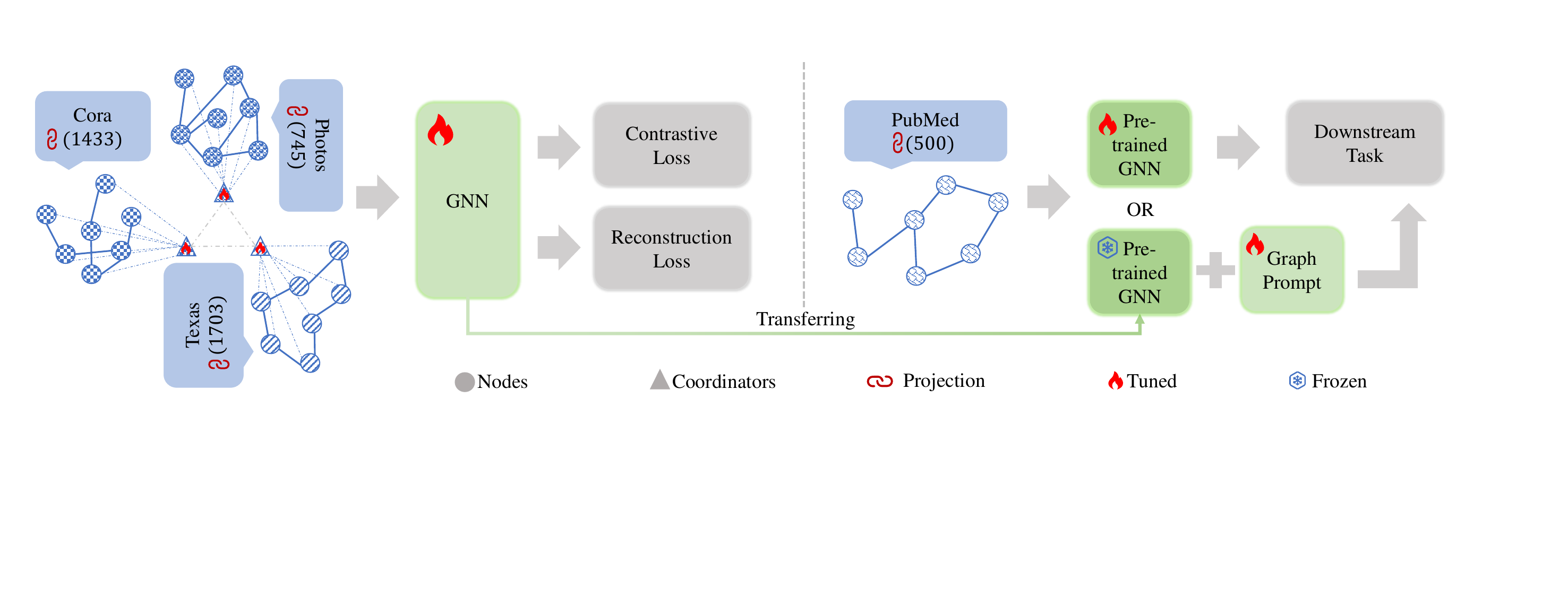}
    \caption{Overview of our proposed GCOPE method. The left part is our pretraining stage and the right part transferring stage.}
    \label{fig:overview}
\end{figure*}

Unlike images, words, or sentences, which often share extensive underlying semantics across datasets, graph data are more abstract, making it more challenging to discern low-level common semantics. Therefore, the negative transfer phenomenon \cite{zhang2022survey,li2024zerog} is commonly found in the field of graph learning. Here, we focus on the representative C-way-K-shot setting. We validate this phenomenon through extensive experimentation, examining two main scenarios: the transfer of knowledge from a single source dataset, as well as from multiple source datasets.

\textit{\textbf{Observations:}} As depicted in Figure~\ref{fig: negative transfer analysis}, transferring from a single source dataset does indeed negatively affect the target task, confirming our analysis of the distinctiveness of two graph datasets.
In order to overcome this obstacle, it is necessary to expand the scope of the source dataset so that it can offer valuable insights for the downstream task. However,  a brutal combination of source datasets still failed to enhance the performance of the target task. We detail this in section \ref{sec:cross}.

\textit{\textbf{Further Analysis:}}
The first reason causing this negative transfer across domains is that the structural patterns are different,  especially reflected in homophilic and heterophilic datasets. Graph data is characterized by its complex network of connections, where nodes are not isolated but part of an interconnected structure. The relationships among nodes, represented by edges, facilitate the flow of influence and information, making each node both a product and a contributor to its environment. This interconnectivity is central to graph data, resulting in systems where the collective properties surpass those of individual components.
Nonetheless, this characteristic presents challenges for machine learning models that attempt to learn from multiple graph datasets simultaneously. Each dataset is akin to its own distinct ecosystem, governed by its unique topology and rules. Within a given graph, nodes and edges are attuned to specific patterns and linkages that are relevant within that context. When merging different graphs for joint training, the inherent disparity of each graph's structure becomes a barrier, as the datasets naturally exist in isolation from one another and thus the information flow is blocked.

The second reason is the lack of feature alignment across these datasets. Graph attributes are heterogeneous and context-dependent, representing a wide range of abstract concepts and connections. Unlike textual or visual data, which have a common reference framework, graph attributes are highly varied and specific to their domain. Consequently, aligning features from different graphs is a daunting task, as there is no straightforward method to reconcile the disparate languages of each dataset into a unified representation for machine learning models to process.

\textit{\textbf{Objectives:}} 
In this paper, we focus on the above two challenges, the disparity and isolation of graph datasets and the difficulty in aligning their diverse features.
We denote our pretraining datasets as comprising $M$ graphs, represented as $\mathcal{G}^{(i)}=(\mathcal{V}^{(i)}, \mathcal{E}^{(i)}), i\in\{1,2,\cdots,M\}$, where $\mathcal{V}^{(i)}=\{v^{(i)}_{1}, v^{(i)}_{2}, \cdots, v^{(i)}_{|\mathcal{V}_i|}\}$ and $\mathcal{E}^{(i)} = \mathcal{V}^{(i)} \times \mathcal{V}^{(i)}$ denote the sets of nodes and edges, respectively. Each $\mathcal{G}^{(i)}$ is associated with feature matrix $X^{(i)} \in \mathbb{R}^{|\mathcal{V}^{(i)}|\times d_i}$ and adjacency matrix $A^{(i)} \in \mathbb{R}^{|\mathcal{V}^{(i)}|\times|\mathcal{V}^{(i)}|}$. Our objective is to train a GNN $h(\cdot)$ parameterized by $\Theta$, which is capable of encoding knowledge transferable to downstream tasks. The downstream dataset is represented as $\mathcal{G}^{(t)}=(\mathcal{V}^{(t)}, \mathcal{E}^{(t)})$ with the feature matrix $X^{(t)}$ and adajcency matrix $A^{(t)}$.
To address this question, we carefully design a general pretraining scheme that is independent of datasets, network architectures and downstream tasks.
\section{Method}
\vspace{-0.25em}
\subsection{Overview of Our Framework}
\vspace{-0.25em}
In this section, we introduce a cohesive approach that enables the simultaneous pretraining of a graph model on multiple datasets. We utilize established pretraining objectives, namely GraphCL \cite{you2020graphCL} and SimGRACE \cite{xia2022simgrace}, to guide the learning process. Additionally, we implement novel techniques specifically designed to overcome the challenges highlighted in Section \ref{sec:motivation}. A visual representation of our methodology is provided in Figure \ref{fig:overview}.

\vspace{-0.75em}
\subsection{Aligning Graphs by Coordinators}
\vspace{-0.25em}
Different graphs usually have different features and structural patterns. Here we propose a two-phase graph alignment approach. The first step is to make the feature dimensions all the same in format. Then we seek to further learn a latent data alignment strategy by coordinators, which can reformulate graphs w.r.t their structural patterns and semantic patterns.
\vspace{-0.65em}
\subsubsection{Feature Projection}
During the pretraining phase of our GNN model, 
we first present a projecting module to align feature dimension, which is described by:
\vspace{-1.25em}

\begin{align}
    \tilde{X}^{(i)} = \mathrm{Proj}(X^{(i)}) \in \mathbb{R}^{|\mathcal{V}^{(i)}|\times d_{\mathrm{P}}},
    \label{eq: proj}
\end{align}
\vspace{-1.25em}

\noindent where $\mathrm{Proj}(\cdot)$ denotes a certain projection operation and $d_{\mathrm{P}}$ denotes the predefined projected dimension. Without loss of generality, we provide two widely used methods, singular value decomposition (SVD) and attention mechanism, as two representative projection operations in this paper.
However, it is worth mentioning that the mere projection of features onto a common plane does not suffice to address the alignment challenge; additional alignment endeavors are indispensable.
\vspace{-0.65em}
\subsubsection{Graph Coordinators}
Based on the above step, we present a novel concept: virtual nodes, referred to herein as ``coordinators''. These coordinators are strategically crafted to enhance inter-connectivity among disparate graphs, facilitating the alignment of their features and structural conflicts.

\textit{ (i). Cross Connection between Coordinators and Datasets:} To ensure that each graph retains its unique structural properties while still participating in a larger, interconnected system, we assign a dedicated coordinator to each graph dataset. This coordinator is not just a peripheral addition; it becomes an integral component of the graph by forming a fully connected subnetwork with every node within the graph to which it is assigned. This design choice ensures that the coordinator has an immediate, direct line of communication with each node, allowing it to efficiently relay information and coordinate interactions within its respective graph.

\textit{ (ii). Inner Connection within Coordinators:}  However, our ambition extends beyond enhancing individual graph structures; we aim to facilitate a rich tapestry of connections spanning all graphs within our dataset collection. To achieve this, we strategically connect coordinators from different graphs. These inter-coordinator edges act as bridges, enabling the flow of information between disparate graphs that would otherwise remain siloed. As a result, we construct a comprehensive communication infrastructure that seamlessly integrates the entirety of the graph collection, fostering a new level of collaboration and knowledge sharing among them. With the above modification, we can get a joint adjacency matrix $\tilde{A}$, which consists of adjacency matrices of original graphs and new rows for coordinators. Mathematically, $\tilde{A}$ can be defined as follows:
\begin{align}
\tilde A= \begin{bmatrix}
A_{\mathrm{diag}} & R_A^T \\
R_A & R_R
\end{bmatrix},
\label{eq:adjacency_after_coordinators}
\end{align}
where $A_{\mathrm{diag}} = \mathrm{Diag}(A^{(1)},A^{(2)},\cdots,A^{(M)})$, $R_R=\mathbf{1}^{M\times M}$, $R_A=\mathrm{Stack}(R_A^{(1)},R_A^{(2)},\cdots,R_A^{(M)})$. $\mathrm{Diag}$ means concatenating matrices diagonally, and $\mathrm{Stack}$ means stacking row-vectors into a matrix. Specifically, $R_A^{(i)}\in \mathbb{R}^{\sum_k^M|\mathcal{V}^{(k)}|}$ and we have the $j$th value of $R_A^{(i)}$:
\begin{align}
R_{A}^{(i)}(j) = 
\begin{cases}
1 & \sum_1^i|\mathcal{V}^{(k)}| \le j < \sum_1^{i+1}|\mathcal{V}^{(k)}| \\
0 & \mathrm{otherwise}.
\end{cases}
\end{align}

\noindent We treat the coordinator representation as learnable parameters. We opt for a flexible approach so that they can evolve organically alongside the GNN training, adapting dynamically to the data. This ensures they continually refine their attributes to effectively serve as conduits of evolving graph embeddings.

\textit{ (iii). Generate Graph Batches for Efficient Training:} Moreover, by harnessing the interconnectivity facilitated by coordinators, we unlock the potential for joint sampling of nodes originating from diverse data graphs. This innovative strategy empowers the training process through aggregated batches, thereby promoting the natural alignment of features across datasets. Exposing the model to a plethora of features within a single learning iteration prompts it to seek out unified representations, effectively synthesizing insights from disparate sources into cohesive and comprehensive representations. This unified approach not only enhances the model's ability to capture the underlying structure of the data but also fosters robust generalization across domains.

\subsubsection{Why It Works?} 
Kindly note that our proposed coordinators have the same theory support as graph prompts. The main difference is that general graph prompts are used for downstream tasks while we use coordinators during pretraining. Evidenced by \cite{fang2022prompt, sun2023all}, prompts are demonstrated to be tantamount to graph transformations (e.g., node, edge, graph removing or adding, and some node feature operations). In parallel, our coordinators can be construed as a form of graph transformation. Given a GNN denoted as $g(\cdot)$ and a prompt $p$, it has been established that:
\begin{align}
g(\tilde{X} + p,\tilde{A}) = g(t(\tilde{X},\tilde{A})) + O_{gp}
\end{align}
where $O_{gp}$ represents an error bound associated with the GNN $g(\cdot)$ and the prompt $p$. Here, $t(\cdot)$ signifies a graph transformation operation. Given that the learnable features embedded within our coordinators can be interpreted as the prompt $p$, they facilitate equivalent graph transformations for the originally isolated graphs. The learned equivalent transformation effectively harmonizes these graphs (\textit{e.g.,} homophilic and heterophilic graphs), enabling the discovery of inherent commonalities among them.


Additionally, we would like to facilitate a better understanding of our framework's design principles and the functions of its various components through `Interdisciplinary Teaching', which is a cutting-edge approach in the field of education:

`Interdisciplinary Teaching' aims to enhance the comprehensiveness and depth of learning by integrating knowledge and methods from different subjects, similar to how GCOPE integrates diverse graph datasets. It addresses the complexity of real-world problems (like diverse and complex structures, attributes, and structure-attribute patterns), requiring a multidisciplinary perspective for solutions and encouraging students to apply cross-domain knowledge. This approach leads to a more thorough understanding and higher-level thinking skills, mostly like the transferrable pretrained graph representations. The integration strategy in education, which highlights core concepts and fosters interaction among disciplines, resembles the coordinators in GCOPE. This helps students see the bridges between subjects, nurturing flexible and innovative thinking, like transferring to new downstream datasets, especially the few-shot scenarios.

\vspace{-0.75em}
\subsection{Pretraining on Multi-domain Graphs}
\vspace{-0.25em}
Our approach is very flexible and compatible with many pretraining approaches. Here, we present a general pretraining framework that can extract high-quality embeddings at both the node and graph levels. Prior research has predominantly focused on pretraining within the same data domains as the downstream task \cite{jin2020self}. Within this context, GraphCL \cite{you2020graphCL} and SimGRACE \cite{xia2022simgrace} have been particularly noteworthy for their effectiveness in generating granular node embeddings as well as holistic graph embeddings. GraphCL employs graph data augmentation to generate positive pairs, whereas SimGRACE perturbs the GNN to achieve this objective. Given the strengths of these methods, we chose them as our basic pretraining strategy.

Next, we move on to the crucial task of preserving the integrity of information from each graph. To this end, we introduce an auxiliary feature reconstruction loss. This loss is quantified through the mean squared error (MSE) metric, which assesses the discrepancy between the original node feature vector (after projection) $\tilde{x}_i$ and the reconstructed feature vector $\hat{x}_i$. The latter is generated by applying a multilayer perceptron (MLP) to the embedded node representation $h_i$, thus enabling the evaluation of the fidelity with which the embedding process preserves the original node features. The purpose of this loss function is to ensure that, despite the projection of features, the salient and defining characteristics of each graph's information are retained and remain compatible. To be concrete, taking GraphCL as an example, our pretraining objective can be described as follows:
\begin{equation}
\vspace{-0.4em}
\footnotesize
\mathcal{L} \!\!=\! -\log\frac{\exp(\mathrm{sim}(h(\mathrm{PS}(\tilde{X},\tilde{A},a_i)),h(\mathrm{PS}(\tilde{X},\tilde{A},a_j))/\tau)}{\sum\exp(\mathrm{sim}(h(\mathrm{PS}(\tilde{X},\tilde{A},a_i)),h(\mathrm{NS}(\tilde{X},\tilde{A},a_j))/\tau)} \!+ \!\|\tilde{X}-\hat{X}\|_2 
\end{equation}
where $\tilde{X}$ is the concatenated feature matrix of all pretraining datasets after projection (Equation \ref{eq: proj}), $\tilde{A}$ is the adjacency matrix after being connected by the coordinators (Equation \ref{eq:adjacency_after_coordinators}), PS, NS respectively denote the positive sampling and negative sampling, sim denotes a similarity metric (\textit{e.g.,} cosine similarity), $a_i, a_j$ are different graph augmentations, and $\lambda$ is the reconstruction loss coefficient that controls how much attention is paid to the reconstruction task.


\subsection{Applying Knowledge to Downstream Data}

Our pretraining methodology is inherently flexible and agnostic to downstream task constraints, making it compatible with various adaptation techniques. For instance, the recent introduction of prompting in the graph field has attracted considerable interest due to its remarkable efficiency and effectiveness \cite{sun2023all}.
To show the superior generalization ability of our GCOPE method, we explore its transferability within both the conventional finetuning paradigm and the nascent graph prompt framework.

Drawing inspiration from the methodology presented by \cite{sun2023all}, we cast downstream tasks into a graph-level framework. As delineated in their work, the maximal retention of pretraining knowledge is observed when downstream tasks are aligned within the same task space as the pretraining task. Given our pretraining is conducted at the graph level, we correspondingly transform all downstream tasks to this level by constructing induced subgraphs. The main algorithm is presented in Algorithm \ref{algo:main}.

\begin{algorithm}[htb]
    \caption{GCOPE}
    \label{algo:main}
    \KwIn{
    Source graphs $\{\mathcal{G}^{(i)}\}_{i=1}^{M}$, target graph $\mathcal{G}^{(t)}$, GNN parameters $\Theta$, projection operation $\mathrm{Proj}(\cdot)$, pretraining objective $\mathcal{L}(\cdot)$, learning rate $\alpha$, transferring pipeline $\mathrm{Trans}(\cdot)$
    }
    \KwOut{
    The optimal model on the target graph $g_t(\cdot)$
    }
    \For{$i\gets0$ \KwTo $M$}
    {$\tilde{X}^{(i)} = \mathrm{Proj}(X^{(i)})$}
    $\tilde{X} = \mathrm{Cat}(\tilde{X}^{(1)},\tilde{X}^{(2)},\cdots,\tilde{X}^{(M)})$ \\
    $\tilde A= \begin{bmatrix}
    A_{\mathrm{diag}} & R_A^T \\
    R_A & R_R
    \end{bmatrix}$ \\
    \While{not converge}
    {$\Theta \gets \Theta - \alpha\nabla_{\Theta}\mathcal{L}(\tilde{X}, \tilde{A}, \Theta)$} 
    $g_t(\cdot) = \mathrm{Trans}(\mathcal{G}^{(t)}, \Theta)$ \\
    \Return{$g_t(\cdot)$}
\end{algorithm}

\vspace{-1.5em}
\subsection{Complexity Analysis} 
Thanks to the simplicity of our pretraining framework, the increase in parameter complexity is only marginal. The only additional parameters introduced are the features of the coordinators, with a complexity of $\mathcal{O}(Md_{\mathrm{A}})$, which scales linearly with the number of pretraining datasets. In practical settings, scenarios characterized by an excessively abundant availability of pretraining datasets are rare. Assuming GNN we employ comprises $L$ layers with a maximum layer width of $d$, and let $N=\sum_{k=1}^{M}|\mathcal{V}^{(k)}|$ and $E=\sum_{k=1}^{M}|\mathcal{E}^{(k)}|$. It is worth noting that the time complexity of a typical graph model, such as Graph Convolutional Network (GCN), is $\mathcal{O}(LNd^2+LEd+Nd)$ \cite{sun2023all}. With the incorporation of coordinators, the revised time complexity becomes $\mathcal{O}(L(N+M)d^2+L(E+N+M)d+(N+M)d)$. The additional time complexity is $\mathcal{O}(LMd^2+L(N+M)d+Md)$. Given that $M\ll N$, the supplementary time cost scales nearly linearly with the number of original nodes.

\begin{table}[tb]
\caption{Statistics of datasets.}
\label{tab:dataset}
\vspace{-1em}
\centering
\resizebox{0.49\textwidth}{!}{
\begin{tabular}{cccccc}
\toprule
Homophilic Data& {Cora} & {Citeseer} & {Pubmed} & {Computers} & {Photos} \\
\midrule
\# Nodes    & 2,708  & 3,327  & 19,717  & 13,752  & 7,650 \\
\# Edges    & 10,556 & 9,104  & 88,648  & 491,722 & 238,162 \\
\# Features & 1,433  & 3,703  & 500     & 767     & 745  \\
\# Labels   & 7      & 6      & 3       & 10      & 8  \\
\( h(G) \)  & 0.810  & 0.736  & 0.802   & 0.777   & 0.827  \\\midrule
Heterophilic Data&{Wisconsin} & {Texas} & {Cornell} & {Chameleon} & {Squirrel}\\\midrule
\# Nodes     & 251    & 183    & 183   & 2,277  & 5,201 \\
\# Edges    & 515    & 325    & 298   & 62792  & 396,846 \\
\# Features & 1,703  & 1,703  & 1,703 & 2,325  & 2,089 \\
\# Labels   & 5      & 5      & 5     & 5      & 5 \\
\( h(G) \)  & 0.196  & 0.108  & 0.305 & 0.231  & 0.222 \\
\bottomrule
\end{tabular}
}
\vspace{-2em}
\end{table}

\begin{table*}[t]
\renewcommand\arraystretch{1.2}
\caption{Cross-domain transfer learning performance (mean±std Acc/AUC/F1) on homophilic datasets (C-way-1-shot). IMP (\%): the average improvement of GCOPE over the rest. GCL and Sim respectively represent GraphCL and SimGRACE.}
\label{tab:rq1 homo}
\vspace{-1em}
\resizebox{1.0\textwidth}{!}{
\begin{tabular}{cc|ccc|ccc|ccc|ccc|ccc}
\cmidrule[1.2pt]{1-17}
\multirow{2}{*}{\begin{tabular}[c]{@{}c@{}}Training\\ schemes\end{tabular}}                   & \multirow{2}{*}{Methods} & \multicolumn{3}{c|}{Cora} & \multicolumn{3}{c|}{Citeseer} & \multicolumn{3}{c|}{Pubmed} & \multicolumn{3}{c|}{Computers} & \multicolumn{3}{c}{Photos} \\
                                                                                              &                          & Acc     & AUC    & F1     & Acc      & AUC      & F1      & Acc     & AUC     & F1      & Acc      & AUC      & F1       & Acc     & AUC     & F1     \\ \midrule
\multirow{4}{*}{supervised}                                                                   & GCN                      & 0.3012\text{\scriptsize ±.06} & 0.6444\text{\scriptsize ±.04} & 0.2591\text{\scriptsize ±.04} & 0.4358\text{\scriptsize ±.09} & 0.7234\text{\scriptsize ±.07} & 0.3583\text{\scriptsize ±.10} & 0.4210\text{\scriptsize ±.01} & 0.6040\text{\scriptsize ±.06} & 0.3026\text{\scriptsize ±.04} & 0.2602\text{\scriptsize ±.07} & 0.6773\text{\scriptsize ±.02} & 0.2428\text{\scriptsize ±.04} & 0.4603\text{\scriptsize ±.04} & 0.8458\text{\scriptsize ±.01} & 0.4592\text{\scriptsize ±.04} \\
                                                                                              & GAT                      & 0.3646\text{\scriptsize ±.04} & 0.6769\text{\scriptsize ±.03} & 0.3108\text{\scriptsize ±.04} & 0.3695\text{\scriptsize ±.05} & 0.7232\text{\scriptsize ±.06} & 0.3305\text{\scriptsize ±.04} & 0.4209\text{\scriptsize ±.04} & 0.5710\text{\scriptsize ±.06} & 0.3227\text{\scriptsize ±.07} & 0.3482\text{\scriptsize ±.07} & 0.6878\text{\scriptsize ±.05} & 0.2397\text{\scriptsize ±.05} & 0.4742\text{\scriptsize ±.08} & 0.8213\text{\scriptsize ±.02} & 0.4498\text{\scriptsize ±.07} \\
                                                                                              & BWGNN                    & 0.2543\text{\scriptsize ±.05} & 0.5563\text{\scriptsize ±.03} & 0.1971\text{\scriptsize ±.02} & 0.3599\text{\scriptsize ±.07} & 0.6954\text{\scriptsize ±.05} & 0.3112\text{\scriptsize ±.06} & 0.3976\text{\scriptsize ±.03} & 0.4934\text{\scriptsize ±.03} & 0.2686\text{\scriptsize ±.04} & 0.2768\text{\scriptsize ±.05} & 0.6273\text{\scriptsize ±.03} & 0.1864\text{\scriptsize ±.03} & 0.4113\text{\scriptsize ±.04} & 0.7769\text{\scriptsize ±.00} & 0.3883\text{\scriptsize ±.01} \\
                                                                                              & FAGCN                    & 0.3819\text{\scriptsize ±.03} & 0.6818\text{\scriptsize ±.04} & 0.3009\text{\scriptsize ±.09} & 0.5219\text{\scriptsize ±.08} & 0.8042\text{\scriptsize ±.03} & 0.4667\text{\scriptsize ±.08} & 0.4522\text{\scriptsize ±.02} & 0.5622\text{\scriptsize ±.04} & 0.4275\text{\scriptsize ±.07} & 0.4651\text{\scriptsize ±.04} & 0.7762\text{\scriptsize ±.02} & 0.3009\text{\scriptsize ±.07} & 0.5937\text{\scriptsize ±.05} & 0.8847\text{\scriptsize ±.00} & 0.5346\text{\scriptsize ±.03} \\ \midrule
\multirow{4}{*}{\begin{tabular}[c]{@{}c@{}}IP \\ + \\ finetuning\end{tabular}} & GCL+GCN            & 0.2507\text{\scriptsize ±.06} & 0.6350\text{\scriptsize ±.03} & 0.2240\text{\scriptsize ±.03} & 0.3140\text{\scriptsize ±.02} & 0.6661\text{\scriptsize ±.04} & 0.2397\text{\scriptsize ±.02} & 0.4217\text{\scriptsize ±.02} & 0.5257\text{\scriptsize ±.05} & 0.2896\text{\scriptsize ±.07} & 0.2856\text{\scriptsize ±.04} & 0.6467\text{\scriptsize ±.03} & 0.1653\text{\scriptsize ±.06} & 0.5533\text{\scriptsize ±.01} & 0.8661\text{\scriptsize ±.01} & 0.5217\text{\scriptsize ±.01} \\
                                                                                              & GCL+FAGCN            & 0.3892\text{\scriptsize ±.05} & 0.7228\text{\scriptsize ±.03} & 0.3619\text{\scriptsize ±.05} & 0.4461\text{\scriptsize ±.02} & 0.7781\text{\scriptsize ±.01} & 0.4126\text{\scriptsize ±.02} & 0.4532\text{\scriptsize ±.02} & 0.5708\text{\scriptsize ±.03} & 0.4168\text{\scriptsize ±.04} & 0.4371\text{\scriptsize ±.06} & 0.7616\text{\scriptsize ±.01} & 0.3450\text{\scriptsize ±.02} & 0.6273\text{\scriptsize ±.01} & 0.8710\text{\scriptsize ±.01} & 0.5406\text{\scriptsize ±.03} \\
                                                                                              & Sim+GCN           & 0.2492\text{\scriptsize ±.02} & 0.5765\text{\scriptsize ±.03} & 0.1567\text{\scriptsize ±.04} & 0.2950\text{\scriptsize ±.06} & 0.6203\text{\scriptsize ±.06} & 0.1812\text{\scriptsize ±.06} & 0.3980\text{\scriptsize ±.01} & 0.5067\text{\scriptsize ±.02} & 0.2805\text{\scriptsize ±.01} & 0.2666\text{\scriptsize ±.10} & 0.6286\text{\scriptsize ±.01} & 0.1603\text{\scriptsize ±.03} & 0.4290\text{\scriptsize ±.04} & 0.7645\text{\scriptsize ±.02} & 0.3955\text{\scriptsize ±.02} \\
                                                                                              & Sim+FAGCN           & 0.3957\text{\scriptsize ±.03} & 0.7284\text{\scriptsize ±.02} & 0.3585\text{\scriptsize ±.01} & 0.5101\text{\scriptsize ±.03} & 0.7969\text{\scriptsize ±.01} & 0.4615\text{\scriptsize ±.04} & 0.4398\text{\scriptsize ±.01} & 0.5535\text{\scriptsize ±.01} & 0.4225\text{\scriptsize ±.02} & 0.4393\text{\scriptsize ±.01} & 0.7718\text{\scriptsize ±.02} & 0.3100\text{\scriptsize ±.02} & 0.5704\text{\scriptsize ±.02} & 0.8543\text{\scriptsize ±.02} & 0.4984\text{\scriptsize ±.01} \\ \midrule
\multirow{4}{*}{\begin{tabular}[c]{@{}c@{}}GCOPE\\ +\\ finetuning\end{tabular}}             & GCL+GCN            & 0.3368\text{\scriptsize ±.02} & 0.6971\text{\scriptsize ±.04} & 0.2967\text{\scriptsize ±.03} & 0.3701\text{\scriptsize ±.03} & 0.7066\text{\scriptsize ±.02} & 0.3265\text{\scriptsize ±.05} & 0.4443\text{\scriptsize ±.04} & 0.5888\text{\scriptsize ±.04} & 0.4242\text{\scriptsize ±.04} & 0.3439\text{\scriptsize ±.03} & 0.7023\text{\scriptsize ±.01} & 0.2976\text{\scriptsize ±.03} & 0.5635\text{\scriptsize ±.02} & 0.8733\text{\scriptsize ±.00} & 0.5480\text{\scriptsize ±.02} \\
                                                                                              & GCL+FAGCN            & 0.4618\text{\scriptsize ±.03} & 0.7597\text{\scriptsize ±.05} & 0.4388\text{\scriptsize ±.05} & 0.5631\text{\scriptsize ±.03} & 0.8258\text{\scriptsize ±.02} & 0.4953\text{\scriptsize ±.04} & 0.4591\text{\scriptsize ±.01} & 0.5512\text{\scriptsize ±.01} & 0.4203\text{\scriptsize ±.03} & 0.4465\text{\scriptsize ±.01} & 0.7747\text{\scriptsize ±.00} & 0.3432\text{\scriptsize ±.03} & 0.6329\text{\scriptsize ±.02} & 0.8850\text{\scriptsize ±.00} & 0.5935\text{\scriptsize ±.03} \\
                                                                                              & Sim+GCN           & 0.2525\text{\scriptsize ±.05} & 0.5744\text{\scriptsize ±.03} & 0.1722\text{\scriptsize ±.06} & 0.3475\text{\scriptsize ±.05} & 0.6527\text{\scriptsize ±.05} & 0.2704\text{\scriptsize ±.05} & 0.4116\text{\scriptsize ±.00} & 0.5166\text{\scriptsize ±.04} & 0.2994\text{\scriptsize ±.03} & 0.3230\text{\scriptsize ±.01} & 0.6994\text{\scriptsize ±.00} & 0.2515\text{\scriptsize ±.00} & 0.4772\text{\scriptsize ±.03} & 0.7851\text{\scriptsize ±.01} & 0.4277\text{\scriptsize ±.02} \\
                                                                                              & Sim+FAGCN           & 0.3875\text{\scriptsize ±.04} & 0.7163\text{\scriptsize ±.03} & 0.3355\text{\scriptsize ±.08} & 0.5704\text{\scriptsize ±.04} & 0.8425\text{\scriptsize ±.01} & 0.5178\text{\scriptsize ±.04} & 0.4727\text{\scriptsize ±.03} & 0.5587\text{\scriptsize ±.03} & 0.5672\text{\scriptsize ±.03} & 0.4677\text{\scriptsize ±.04} & 0.7875\text{\scriptsize ±.01} & 0.3823\text{\scriptsize ±.02} & 0.5985\text{\scriptsize ±.02} & 0.8757\text{\scriptsize ±.02} & 0.5556\text{\scriptsize ±.05} \\ \midrule
\multicolumn{2}{c|}{IMP (\%)}                                                                                             & 11.23\%            & 5.23\%             & 14.63\%            & 13.81\%            & 4.26\%            & 16.59\%            & 5.02\%            & 0.99\%            & 25.32\%            & 13.79\%            & 6.28\%            & 30.70\%            & 10.31\%            & 2.30\%            & 12.18\%            \\
\cmidrule[1.2pt]{1-17}
\end{tabular}
}
\vspace{-1em}
\end{table*}

\begin{table*}[t]
\renewcommand\arraystretch{1.2}
\caption{Cross-domain transfer learning performance (mean±std Acc/AUC/F1) on heterophilic datasets (C-way-1-shot).}
\label{tab:rq1 hetero}
\vspace{-1em}
\resizebox{1.0\textwidth}{!}{
\begin{tabular}{cc|ccc|ccc|ccc|ccc|ccc}
\cmidrule[1.2pt]{1-17}
\multirow{2}{*}{\begin{tabular}[c]{@{}c@{}}Training\\ schemes\end{tabular}}                   & \multirow{2}{*}{Methods} & \multicolumn{3}{c|}{Wisconsin} & \multicolumn{3}{c|}{Texas} & \multicolumn{3}{c|}{Cornell} & \multicolumn{3}{c|}{Chameleon} & \multicolumn{3}{c}{Squirrel} \\
                                                                                              &                          & Acc      & AUC      & F1       & Acc     & AUC     & F1     & Acc      & AUC     & F1      & Acc      & AUC      & F1       & Acc      & AUC      & F1      \\ \midrule
\multirow{4}{*}{supervised}                                                                   & GCN                      & 0.6290\text{\scriptsize ±.05} & 0.8320\text{\scriptsize ±.04} & 0.4871\text{\scriptsize ±.14} & 0.5812\text{\scriptsize ±.08} & 0.6731\text{\scriptsize ±.04} & 0.4557\text{\scriptsize ±.10} & 0.3263\text{\scriptsize ±.04} & 0.5666\text{\scriptsize ±.01} & 0.3151\text{\scriptsize ±.03} & 0.2393\text{\scriptsize ±.03} & 0.5310\text{\scriptsize ±.04} & 0.1923\text{\scriptsize ±.03} & 0.2093\text{\scriptsize ±.00} & 0.5263\text{\scriptsize ±.01} & 0.1889\text{\scriptsize ±.01} \\
                                                                                              & GAT                      & 0.6009\text{\scriptsize ±.02} & 0.8346\text{\scriptsize ±.01} & 0.5217\text{\scriptsize ±.05} & 0.6300\text{\scriptsize ±.08} & 0.5854\text{\scriptsize ±.08} & 0.4282\text{\scriptsize ±.13} & 0.3275\text{\scriptsize ±.14} & 0.5306\text{\scriptsize ±.03} & 0.1497\text{\scriptsize ±.04} & 0.2342\text{\scriptsize ±.02} & 0.5205\text{\scriptsize ±.04} & 0.1379\text{\scriptsize ±.03} & 0.2118\text{\scriptsize ±.00} & 0.5195\text{\scriptsize ±.02} & 0.1160\text{\scriptsize ±.01} \\
                                                                                              & BWGNN                    & 0.5620\text{\scriptsize ±.05} & 0.8463\text{\scriptsize ±.02} & 0.5189\text{\scriptsize ±.05} & 0.7438\text{\scriptsize ±.10} & 0.6642\text{\scriptsize ±.07} & 0.6274\text{\scriptsize ±.22} & 0.3150\text{\scriptsize ±.09} & 0.5938\text{\scriptsize ±.06} & 0.2190\text{\scriptsize ±.05} & 0.2206\text{\scriptsize ±.02} & 0.5039\text{\scriptsize ±.03} & 0.1540\text{\scriptsize ±.03} & 0.2155\text{\scriptsize ±.00} & 0.5149\text{\scriptsize ±.00} & 0.1664\text{\scriptsize ±.02} \\
                                                                                              & FAGCN                    & 0.5222\text{\scriptsize ±.05} & 0.7905\text{\scriptsize ±.0310} & 0.4725\text{\scriptsize ±.06} & 0.6900\text{\scriptsize ±.06} & 0.7185\text{\scriptsize ±.01} & 0.5334\text{\scriptsize ±.12} & 0.2938\text{\scriptsize ±.06} & 0.6573\text{\scriptsize ±.04} & 0.2872\text{\scriptsize ±.05} & 0.2575\text{\scriptsize ±.02} & 0.5515\text{\scriptsize ±.02} & 0.1941\text{\scriptsize ±.01} & 0.2181\text{\scriptsize ±.00} & 0.5202\text{\scriptsize ±.00} & 0.1875\text{\scriptsize ±.02} \\ \midrule
\multirow{4}{*}{\begin{tabular}[c]{@{}c@{}}IP \\ +\\ finetuning\end{tabular}} & GCL+GCN            & 0.5249\text{\scriptsize ±.03} & 0.7876\text{\scriptsize ±.03} & 0.4415\text{\scriptsize ±.05} & 0.7350\text{\scriptsize ±.01} & 0.7210\text{\scriptsize ±.02} & 0.5636\text{\scriptsize ±.09} & 0.4175\text{\scriptsize ±.04} & 0.6350\text{\scriptsize ±.02} & 0.3500\text{\scriptsize ±.04} & 0.2249\text{\scriptsize ±.02} & 0.5213\text{\scriptsize ±.00} & 0.1432\text{\scriptsize ±.03} & 0.2118\text{\scriptsize ±.01} & 0.5059\text{\scriptsize ±.01} & 0.1110\text{\scriptsize ±.03} \\
                                                                                              & GCL+FAGCN            & 0.6063\text{\scriptsize ±.04} & 0.8356\text{\scriptsize ±.01} & 0.5555\text{\scriptsize ±.07} & 0.7425\text{\scriptsize ±.03} & 0.7034\text{\scriptsize ±.03} & 0.6141\text{\scriptsize ±.09} & 0.2588\text{\scriptsize ±.04} & 0.6262\text{\scriptsize ±.04} & 0.2442\text{\scriptsize ±.04} & 0.2443\text{\scriptsize ±.00} & 0.5530\text{\scriptsize ±.01} & 0.1875\text{\scriptsize ±.01} & 0.2223\text{\scriptsize ±.00} & 0.5307\text{\scriptsize ±.00} & 0.1740\text{\scriptsize ±.02} \\
                                                                                              & Sim+GCN           & 0.5258\text{\scriptsize ±.04} & 0.7927\text{\scriptsize ±.05} & 0.4604\text{\scriptsize ±.06} & 0.6338\text{\scriptsize ±.05} & 0.6024\text{\scriptsize ±.07} & 0.4269\text{\scriptsize ±.14} & 0.3438\text{\scriptsize ±.13} & 0.5954\text{\scriptsize ±.09} & 0.2168\text{\scriptsize ±.09} & 0.2271\text{\scriptsize ±.01} & 0.5183\text{\scriptsize ±.02} & 0.1578\text{\scriptsize ±.03} & 0.2133\text{\scriptsize ±.00} & 0.5133\text{\scriptsize ±.01} & 0.1550\text{\scriptsize ±.02} \\
                                                                                              & Sim+FAGCN           & 0.6335\text{\scriptsize ±.02} & 0.8557\text{\scriptsize ±.00} & 0.5830\text{\scriptsize ±.04} & 0.6725\text{\scriptsize ±.14} & 0.6922\text{\scriptsize ±.04} & 0.5906\text{\scriptsize ±.10} & 0.2725\text{\scriptsize ±.05} & 0.6433\text{\scriptsize ±.04} & 0.2617\text{\scriptsize ±.04} & 0.2748\text{\scriptsize ±.01} & 0.5652\text{\scriptsize ±.00} & 0.2011\text{\scriptsize ±.00} & 0.2170\text{\scriptsize ±.00} & 0.5213\text{\scriptsize ±.00} & 0.1716\text{\scriptsize ±.01} \\ \midrule
\multirow{4}{*}{\begin{tabular}[c]{@{}c@{}}GCOPE\\ +\\ finetuning\end{tabular}}             & GCL+GCN            & 0.6606\text{\scriptsize ±.04} & 0.8487\text{\scriptsize ±.01} & 0.5952\text{\scriptsize ±.04} & 0.7738\text{\scriptsize ±.06} & 0.7387\text{\scriptsize ±.01} & 0.6763\text{\scriptsize ±.08} & 0.3975\text{\scriptsize ±.10} & 0.6694\text{\scriptsize ±.04} & 0.3120\text{\scriptsize ±.04} & 0.2411\text{\scriptsize ±.01} & 0.5564\text{\scriptsize ±.00} & 0.2210\text{\scriptsize ±.00} & 0.2245\text{\scriptsize ±.00} & 0.5207\text{\scriptsize ±.01} & 0.1741\text{\scriptsize ±.00} \\
                                                                                              & GCL+FAGCN            & 0.6579\text{\scriptsize ±.03} & 0.8531\text{\scriptsize ±.01} & 0.5649\text{\scriptsize ±.00} & 0.7125\text{\scriptsize ±.02} & 0.6693\text{\scriptsize ±.02} & 0.6300\text{\scriptsize ±.03} & 0.4013\text{\scriptsize ±.05} & 0.6897\text{\scriptsize ±.01} & 0.3160\text{\scriptsize ±.02} & 0.2886\text{\scriptsize ±.00} & 0.5898\text{\scriptsize ±.00} & 0.2320\text{\scriptsize ±.00} & 0.2257\text{\scriptsize ±.00} & 0.5257\text{\scriptsize ±.00} & 0.1885\text{\scriptsize ±.01} \\
                                                                                              & Sim+GCN           & 0.5412\text{\scriptsize ±.03} & 0.8059\text{\scriptsize ±.02} & 0.4509\text{\scriptsize ±.06} & 0.6137\text{\scriptsize ±.18} & 0.6900\text{\scriptsize ±.03} & 0.4674\text{\scriptsize ±.10} & 0.3675\text{\scriptsize ±.09} & 0.6045\text{\scriptsize ±.04} & 0.2339\text{\scriptsize ±.04} & 0.2573\text{\scriptsize ±.02} & 0.5467\text{\scriptsize ±.01} & 0.1852\text{\scriptsize ±.01} & 0.2180\text{\scriptsize ±.00} & 0.5147\text{\scriptsize ±.00} & 0.1783\text{\scriptsize ±.00} \\
                                                                                              & Sim+FAGCN           & 0.7321\text{\scriptsize ±.00} & 0.9305\text{\scriptsize ±.00} & 0.6873\text{\scriptsize ±.01} & 0.7950\text{\scriptsize ±.03} & 0.7451\text{\scriptsize ±.01} & 0.7042\text{\scriptsize ±.03} & 0.5925\text{\scriptsize ±.01} & 0.8069\text{\scriptsize ±.03} & 0.4626\text{\scriptsize ±.03} & 0.2894\text{\scriptsize ±.01} & 0.5662\text{\scriptsize ±.02} & 0.2192\text{\scriptsize ±.02} & 0.2193\text{\scriptsize ±.00} & 0.5370\text{\scriptsize ±.00} & 0.1984\text{\scriptsize ±.01} \\ \midrule
\multicolumn{2}{c|}{IMP~(\%)}                                                                                             & 12.57\%            & 4.58\%            & 13.76\%            & 6.65\%            & 6.08\%            & 16.87\%            & 37.66\%            & 14.29\%            & 29.62\%            & 11.97\%            & 6.01\%            & 25.36\%            & 3.25\%            & 1.06\%            & 16.39\%            \\
\cmidrule[1.2pt]{1-17}
\end{tabular}
}
\vspace{-1.25em}
\end{table*}

\vspace{-0.5em}
\section{Experiments}
\vspace{-0.2em}
In this section, we study the experimental results of our proposed method and baselines to answer five research questions:

\begin{itemize}[leftmargin=1.5em]
    \item \textbf{RQ1.} What advantages does our proposed GCOPE method offer over representative baselines in cross-domain transfer learning?
    \item \textbf{RQ2.} How significant is the presence of edges among different coordinators?
    \item \textbf{RQ3.} Is the reconstruction loss necessary in GCOPE? 
    \item \textbf{RQ4.} Can downstream prompts derive benefits from the cross-domain pretraining?
    \item \textbf{RQ5.} Which projection operation is more compatible for CCOPE?

\end{itemize}

\vspace{-1em}
\subsection{Experimental Setup}

\vspace{-0.25em}
\paragraph{\textbf{Datasets.}} For comprehensive comparisons, we conduct experiments on ten real-world datasets. We choose five homophily datasets in experiments, including Cora \cite{sen2008homoCoraCiteseer}, Citeseer \cite{sen2008homoCoraCiteseer},  Pubmed \cite{namata2012homoPubmed}, Computers and Photos datasets \cite{mcauley2015homoComputerPhoto,shchur2018homoComputersPhoto}, and five heterophilic datasets, including  three sub-datasets derived from the WebKB~\cite{pei2019heterophilicDatasets} (Cornell, Texas, and Wisconsin) and two page-page networks (Chameleon and Squirrel) extracted from Wikipedia \cite{pei2019heterophilicDatasets}. Table~\ref{tab:dataset} summarizes the details, where $h(G)$ is a metric that represents the degrees of homophily and heterophily. The values of $h(G)$ are directly drawn from~\cite{xu2023homoheterophily}, indicating that the first five datasets are highly homophilic and the latter five are highly heterophilic~\cite{xu2023homoheterophily,luan2023HomoHetero}. Different degrees of homophily and heterophily indicate different semantics in graphs.  For more detailed information on these datasets, please check in Appendix.

\vspace{-0.25em}
\paragraph{\textbf{Baselines.}} We compare our proposed method with the following baselines, categorized into three groups and accompanied by concise descriptions. \textbf{(1) Supervised methods:} These methods train a GNN model on a downstream task and infer results directly. In this work, four famous GNN models are adopted, including GCN~\cite{kipf2017GCN}, GAT~\cite{velivckovic2018GAT}, BWGNN~\cite{tang2022BWGNN}, and FAGCN~\cite{bo2021FAGCN}. 
We choose GCN and FAGCN as the backbones for our proposed GCOPE method since FAGCN is tailored to address both homophilic and heterophilic graphs~\cite{xu2023homoheterophily} and GCN, a widely used GNN model, is the basis of FAGCN.
\textbf{(2) Isolated Pretraining (IP) with finetuning:} 
These methods initially utilize multiple cross-domain datasets as source datasets, which are combined in an isolated manner to pretrain a GNN model in a self-supervised fashion (\textit{e.g.} GraphCL~\cite{you2020graphCL} or SimGRACE~\cite{xia2022simgrace}). Here, `isolated' denotes that the datasets are amalgamated into a batch object without establishing inter-dataset connections, resulting in an adjacency matrix composed of distinct blocks. Subsequently, the pretrained model undergoes finetuning for a new downstream task.
\textbf{(3) Graph Coordinators for Pretraining (GCOPE) with transferring:} With learnable coordinators, our proposed GCOPE method aims to unify the isolated source datasets into one large-scale dataset with inter-dataset connections for pretraining and then transfer the pretrained GNN model to a downstream task via finetuning or prompting~\cite{sun2023all}.

\vspace{-0.5em}
\paragraph{\textbf{Metrics and Implementations.}} 
 We choose three widely used metrics to evaluate the node classification task \cite{sun2023all,xu2023homoheterophily,hart2023auc,anonymous2024F1Score}, including classification accuracy (Acc), mean AUC-ROC~(AUC), and mean F1-score~(F1). 
We leverage a 10-fold partition strategy to split the 10 real-world datasets, nine datasets as cross-domain source datasets for pretraining and the rest one dataset as the target dataset for transferring. To unify features of multiple cross-domain source datasets onto one single plane, we leverage SVD to reduce the initial features to 100 dimensions. 
We fix the number of graph neural layers at 2, with a hidden dimension of 100 for GCN, FAGCN, and GAT models. Additionally, we adopt the identical hyperparameters as specified in \cite{tang2022BWGNN} for BWGNN.
For all networks, we apply the Adam optimizer and the learning rate is assigned as $0.0001$. In terms of GCOPE, we introduce one coordinator for each source dataset and assign $0.2$ as the default reconstruction weight. Finally, in the transferring stage, we adopt node classification as the downstream task. For fair comparisons, we apply the C-way-K-shot learning setting, same as~\cite{gprompt23}, for each target dataset to build the training data and then split the rest of the data randomly in 1:9 for validating and testing. In this paper, we report average results on all downstream tasks. More implementation details are shown in Appendix~\ref{appendix: exps}, in which we also analyze the performance of GCOPE with more coordinators and dynamical edges between coordinators.

\vspace{-1em}
\subsection{Cross-domain Transfer Performance with Few-shot Learning Settings (RQ1)}\label{sec:cross}
We compare our proposed GCOPE method with all the baselines on the downstream node classification tasks under the C-way-1-shot setting. We repeat the evaluation 5 times and report the average results and standard deviations in Table~\ref{tab:rq1 homo} and Table~\ref{tab:rq1 hetero}. The results of supervised methods are the benchmark to help validate whether the pretrained GNN models effectively transfer to the downstream tasks. As is known, pretraining GNNs aims to transfer prior knowledge from upstream tasks to improve the performance of GNNs on downstream tasks, especially in few-shot scenarios. However, we observe that most IP with finetuning methods struggle to achieve better performance compared with supervised methods, embodying the negative transfer phenomenon. This is due to the apparent differences in distribution across cross-domain datasets. Under the IP strategy, each graph sample only contains information from one of the nine distributions, making it challenging for GNN models to fit the nine independent distributions into a unified one and learn general graph representations effectively. In contrast, our proposed GCOPE with finetuning methods significantly outperforms almost all baselines, achieving positive transfer. This is because coordinators connect all the source datasets together, unifying nine independent distributions into one cohesive joint distribution. In the pretraining stage, all the training samples are drawn from this joint distribution, facilitating the sharing of information across cross-domain datasets. Consequently, the GNN model is able to balance the shared information and learn better graph representations for transferring knowledge on the downstream task. The reported improvements range from 3.25\% up to 37.66\% in terms of node classification accuracy. Kindly note that our few-shot experiment settings are different from the classic few-shot node classification settings that are actually designed for the meta-learning task. Additionally, we also evaluate our proposed GCOPE on C-way-3-shot and C-way-5-shot settings, shown in Table~\ref{tab:rq1 homo 3 shot} and Table~\ref{tab:rq1 hetero 3 shot}.

\vspace{-1.0em}
\subsection{Interconnectivity Analysis (RQ2)}
\vspace{-0.25em}
In this section, we investigate the impact of edges between different coordinators on the effectiveness of our proposed GCOPE framework. Specifically, we compare two variants: GCOPE/w, which includes inter-coordinator edges, and GCOPE/wo, which excludes them. Both variants share the same hyperparameters, with FAGCN as the backbone model and pretraining based on GraphCL. Additionally, the datasets utilized for pretraining and downstream tasks are partitioned using the 10-fold strategy. Table~\ref{tab:rq3 no cross-link edge} presents the node classification accuracy (mean±std). Our experimental findings underscore the essential and effective role of inter-coordinator edges in GCOPE.

\vspace{-1.0em}
\subsection{Reconstruction Analysis (RQ3)}
\begin{figure*}[htbp]
  \includegraphics[width=1.0\textwidth]{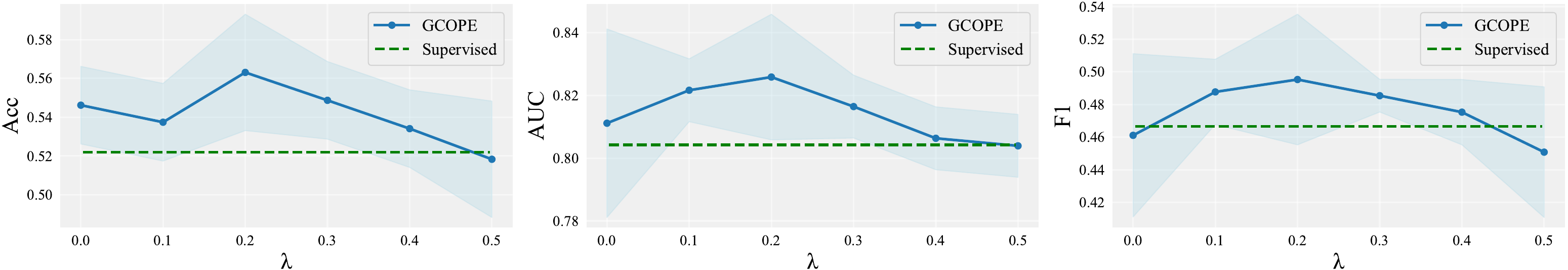}
  \vspace{-2em}
  \caption{Node classification performance (mean±std) of GCOPE with varying reconstruction loss coefficient ($\lambda$) values on Citeseer under C-way-1-shot setting.}
  \label{fig: rq3 hyper analysis}
  \vspace{-0.5em}
\end{figure*}

\vspace{-0.25em}
We conduct a comparison of downstream node classification performance on Citeseer between our proposed GCOPE method with varying reconstruction loss coefficient values $\lambda$, and a supervised method to assess the efficacy of the reconstruction module. Specifically, FAGCN serves as the backbone model for both GCOPE and the supervised method. For GCOPE, GraphCL is utilized as the pretraining strategy. All other hyperparameters between GCOPE and the supervised method are maintained consistent. Figure~\ref{fig: rq3 hyper analysis} depicts the experimental results, focusing on Acc, AUC, and F1 metrics.

Our experimental findings yield the following observations: First, GCOPE without reconstruction ($\lambda=0.0$) outperforms the supervised pretraining, highlighting the effectiveness of the introduced coordinators. Second, GCOPE with reconstruction ($\lambda>0.0$) achieves optimal performance when $\lambda$ is set to 0.2, surpassing both the supervised pretraining and GCOPE ($\lambda=0.0$). This improvement is attributed to the reconstruction module's ability to align graph features across datasets, facilitating more effective learning of shared information by GNNs from cross-domain source datasets. Third, as $\lambda$ exceeds 0.2, performance begins to deteriorate, ultimately falling short of both the supervised pretraining and GCOPE ($\lambda=0.0$). This decline can be attributed to excessively large $\lambda$ values, which cause the model to overly prioritize reconstruction at the expense of its primary pretraining task. In summary, the inclusion of the reconstruction module with relatively small $\lambda$ values proves essential for our proposed GCOPE method.

{
\small
\begin{table}[tbp]
\caption{Node classification accuracy (mean±std) of GCOPE/wo and GCOPE/w, which represent GCOPE without or with inter-coordinator edges respectively.}
\label{tab:rq3 no cross-link edge}
\vspace{-1.5em}
\resizebox{0.48\textwidth}{!}{
\begin{tabular}{c|cc|c|cc}
\cmidrule[1.2pt]{1-6}
Homo Data & GCOPE/wo & GCOPE/w & Hetero Data & GCOPE/wo & GCOPE/w \\ \cmidrule[1.2pt]{1-6}
Cora                                   & 0.3575\text{\scriptsize ±.02}        & 0.4618\text{\scriptsize ±.03}  &  Wisconsin                              & 0.6335\text{\scriptsize ±.04}        & 0.6579\text{\scriptsize ±.03}    \\ \midrule
Citeseer                               & 0.5062\text{\scriptsize ±.02}        & 0.5631\text{\scriptsize ±.03} & Texas                                  & 0.6863\text{\scriptsize ±.04}        & 0.7125\text{\scriptsize ±.02}      \\ \midrule
Pubmed                                 & 0.4108\text{\scriptsize ±.03}        & 0.4591\text{\scriptsize ±.01} & Cornell                                & 0.3212\text{\scriptsize ±.03}        & 0.4013\text{\scriptsize ±.05}      \\ \midrule
Computers                              & 0.4244\text{\scriptsize ±.01}        & 0.4465\text{\scriptsize ±.01} & Chameleon                              & 0.2647\text{\scriptsize ±.01}        & 0.2886\text{\scriptsize ±.00}      \\ \midrule
Photos                                 & 0.6037\text{\scriptsize ±.03}        & 0.6329\text{\scriptsize ±.02} & Squirrel                               & 0.2177\text{\scriptsize ±.00}        & 0.2257\text{\scriptsize ±.00}   \\  \cmidrule[1.2pt]{1-6}
\end{tabular}
}
\vspace{-2.25em}
\end{table}
}

\begin{table*}[htbp]
\renewcommand\arraystretch{1.0}
\caption{Cross-domain transfer learning performance (mean±std Acc/AUC/F1) of GCOPE with ProG (C-way-1-shot). GCL and Sim respectively represent GraphCL and SimGRACE.}
\label{tab:rq4 prompt}
\vspace{-1em}
\resizebox{1.0\textwidth}{!}{
\begin{tabular}{cc|ccc|ccc|ccc|ccc}
\cmidrule[1.2pt]{1-14}
\multirow{2}{*}{\begin{tabular}[c]{@{}c@{}}Training\\ schemes\end{tabular}}     & \multirow{2}{*}{Methods} & \multicolumn{3}{c|}{Cora} & \multicolumn{3}{c|}{Citeseer} & \multicolumn{3}{c|}{Wisconsin} & \multicolumn{3}{c}{Texas}\\
                                                                                &                          & Acc     & AUC    & F1     & Acc      & AUC      & F1      & Acc     & AUC     & F1      & Acc      & AUC      & F1     \\ \midrule
supervised                                                                      & FAGCN                    & 0.3819\text{\scriptsize ±.03} & 0.6818\text{\scriptsize ±.04} & 0.3009\text{\scriptsize ±.09} & 0.5219\text{\scriptsize ±.08} & 0.8042\text{\scriptsize ±.03} & 0.4667\text{\scriptsize ±.08} & 0.5222\text{\scriptsize ±.03} & 0.7905\text{\scriptsize ±.03} & 0.4725\text{\scriptsize ±.06} & 0.6900\text{\scriptsize ±.06} & 0.7185\text{\scriptsize ±.01} & 0.5334\text{\scriptsize ±.01} \\ \midrule
\multirow{2}{*}{\begin{tabular}[c]{@{}c@{}}IP + \\ finetuning\end{tabular}}       & GCL+FAGCN                & 0.3892\text{\scriptsize ±.05} & 0.7228\text{\scriptsize ±.03} & 0.3619\text{\scriptsize ±.05} & 0.4461\text{\scriptsize ±.02} & 0.7781\text{\scriptsize ±.01} & 0.4126\text{\scriptsize ±.02} & 0.6063\text{\scriptsize ±.04} & 0.8356\text{\scriptsize ±.01} & 0.5555\text{\scriptsize ±.07} & 0.7425\text{\scriptsize ±.03} & 0.7034\text{\scriptsize ±.03} & 0.6141\text{\scriptsize ±.00} \\
                                                                                & Sim+FAGCN                & 0.3957\text{\scriptsize ±.03} & 0.7284\text{\scriptsize ±.02} & 0.3858\text{\scriptsize ±.01} & 0.5101\text{\scriptsize ±.03} & 0.7969\text{\scriptsize ±.01} & 0.4615\text{\scriptsize ±.04} & 0.6335\text{\scriptsize ±.02} & 0.8557\text{\scriptsize ±.00} & 0.5830\text{\scriptsize ±.04} & 0.6725\text{\scriptsize ±.01} & 0.6922\text{\scriptsize ±.04} & 0.5906\text{\scriptsize ±.10} \\ 
                                                                                \midrule
\multirow{2}{*}{\begin{tabular}[c]{@{}c@{}}GCOPE + \\ ProG\end{tabular}}       & GCL+FAGCN                & 0.4283\text{\scriptsize ±.03} & 0.7552\text{\scriptsize ±.01} & 0.4326\text{\scriptsize ±.01} & 0.5251\text{\scriptsize ±.01} & 0.8314\text{\scriptsize ±.00} & 0.5042\text{\scriptsize ±.01} & 0.6208\text{\scriptsize ±.04} & 0.8770\text{\scriptsize ±.01} & 0.6359\text{\scriptsize ±.07} & 0.7438\text{\scriptsize ±.01} & 0.6689\text{\scriptsize ±.00} & 0.5880\text{\scriptsize ±.05} \\
                                                                                & Sim+FAGCN                & 0.3941\text{\scriptsize ±.01} & 0.6892\text{\scriptsize ±.03} & 0.3507\text{\scriptsize ±.09} & 0.5270\text{\scriptsize ±.01} & 0.8051\text{\scriptsize ±.01} & 0.4821\text{\scriptsize ±.02} & 0.7312\text{\scriptsize ±.01} & 0.8931\text{\scriptsize ±.00} & 0.6579\text{\scriptsize ±.01} & 0.7700\text{\scriptsize ±.08} & 0.7356\text{\scriptsize ±.01} & 0.7353\text{\scriptsize ±.06} \\ 
                                                                                \midrule
\multirow{2}{*}{\begin{tabular}[c]{@{}c@{}}GCOPE + \\ finetuning\end{tabular}}    & GCL+FAGCN                & 0.4618\text{\scriptsize ±.03} & 0.7597\text{\scriptsize ±.05} & 0.4388\text{\scriptsize ±.05} & 0.5631\text{\scriptsize ±.03} & 0.8258\text{\scriptsize ±.02} & 0.4953\text{\scriptsize ±.04} & 0.6579\text{\scriptsize ±.03} & 0.8531\text{\scriptsize ±.01} & 0.5649\text{\scriptsize ±.00} & 0.7125\text{\scriptsize ±.02} & 0.6693\text{\scriptsize ±.02} & 0.6300\text{\scriptsize ±.03} \\
                                                                                & Sim+FAGCN                & 0.3875\text{\scriptsize ±.04} & 0.7163\text{\scriptsize ±.03} & 0.3355\text{\scriptsize ±.08} & 0.5704\text{\scriptsize ±.04} & 0.8425\text{\scriptsize ±.01} & 0.5178\text{\scriptsize ±.04} & 0.7321\text{\scriptsize ±.00} & 0.9305\text{\scriptsize ±.00} & 0.6873\text{\scriptsize ±.01} & 0.7950\text{\scriptsize ±.03} & 0.7451\text{\scriptsize ±.01} & 0.7042\text{\scriptsize ±.03} \\
\cmidrule[1.2pt]{1-14}
\end{tabular}
}
\end{table*}

\vspace{-1.0em}
\subsection{Transferring by Graph Prompt (RQ4)}
\vspace{-0.25em}
In addition to finetuning, we explore the feasibility of leveraging the graph prompt technique to transfer upstream cross-domain knowledge learned by GCOPE. Specifically, we adopt the implementation of ProG~\cite{sun2023all}, a widely used graph prompt method, which involves freezing the parameters of pretrained GNNs and incorporating graph prompts in the downstream node classification task. Subsequently, we evaluate the performance of GCOPE with ProG on four downstream datasets, comprising two homophilic and two heterophilic datasets. 
The experimental results are presented in Table~\ref{tab:rq4 prompt}.
For comparison, we include results of the supervised method, IP with finetuning, and GCOPE with finetuning.

Based on our experimental observations, we draw the following conclusions: In comparison to the supervised and IP with finetuning methods, both GCOPE with finetuning and GCOPE with ProG exhibit superior performance. Notably, GCOPE with ProG achieves positive transfer with minimal tunable parameters in the downstream node classification task. While the performance of GCOPE with ProG is slightly lower than that of GCOPE with finetuning, the disparity between the two methods is significantly narrower than that between GCOPE with ProG and the supervised method. Based on the aforementioned analysis, we can assert that our proposed GCOPE framework can effectively benefit downstream prompts.


{
\small
\begin{table}[h]
\vspace{-1em}
\caption{Accuracy (mean±std) of GCOPE with finetuning on different projection operations. Experiment settings are the same as Table \ref{tab:rq1 homo}.}
\label{tab:rq5 compatibility analysis}
\vspace{-1em}
\resizebox{0.48\textwidth}{!}{
\begin{tabular}{c|cc|c|cc}
\cmidrule[1.2pt]{1-6}
Homo Data    & Attention                        & SVD                            & Hetero Data   & Attention                        & SVD \\ \cmidrule[1.2pt]{1-6}
Cora         & 0.3120\text{\scriptsize ±.02}    & 0.4618\text{\scriptsize ±.03}  & Wisconsin     & 0.5692\text{\scriptsize ±.05}    & 0.6579\text{\scriptsize ±.03}     \\ \midrule
Citeseer     & 0.3861\text{\scriptsize ±.07}    & 0.5631\text{\scriptsize ±.03}  & Texas         & 0.6663\text{\scriptsize ±.06}    & 0.7125\text{\scriptsize ±.02}     \\ \midrule
Pubmed       & 0.4475\text{\scriptsize ±.02}    & 0.4591\text{\scriptsize ±.01}  & Cornell       & 0.3737\text{\scriptsize ±.03}    & 0.4013\text{\scriptsize ±.05}     \\ \midrule
Computers    & 0.3689\text{\scriptsize ±.06}    & 0.4465\text{\scriptsize ±.01}  & Chameleon     & 0.2755\text{\scriptsize ±.00}    & 0.2886\text{\scriptsize ±.00}     \\ \midrule
Photos       & 0.4857\text{\scriptsize ±.04}    & 0.6329\text{\scriptsize ±.02}  & Squirrel      & 0.2210\text{\scriptsize ±.00}    & 0.2257\text{\scriptsize ±.00}     \\ 
\cmidrule[1.2pt]{1-6}
\end{tabular}
}
\vspace{-2em}
\end{table}
}

\vspace{-1.0em}
\subsection{Compatibility Analysis of Projection Operation (RQ5)}
\vspace{-0.25em}
To study the compatibility of various projection operations, we evaluate our proposed GCOPE method with SVD or attention mechanism across ten cross-domain real-world datasets. Specifically, we set FAGCN as the backbone, leverage GraphCL for pretraining, and finally transfer the pretrained GNNs to the downstream C-way 1-shot node classification tasks. For GCOPE with attention mechanism, we equip each dataset with a data-specific attention module to project their respective features into a shared plane, regardless of their position in the pretraining and finetuning pipeline. We report the node classification accuracy (mean$\pm$std ) in Table~\ref{tab:rq5 compatibility analysis}. We can easily observe that GCOPE with attention mechanism significantly underperforms the ones with SVD. This could be attributed to the fact that the utilized samples are insufficient to train attention modules effectively. Each attention module has $32,792,288$ tuned parameters, while SVD is a non-parametric mathematical method. Therefore, we ensure that SVD is more compatible under few-shot learning scenarios. For effectiveness and simplicity, we set SVD as the default projection operation of GCOPE in this work.

\section{Related Work}
\noindent \textbf{Graph Pretraining.} In the machine learning (ML) research field, pretraining is widely acknowledged for its ability to leverage existing data to train a feature extractor with robust generalization capabilities \cite{devlin2018bert, ridnik2021imagenet, zhao2024kdalign, li2023unlocking, liu2024segno}. Specifically within the graph domain, pretraining methods can be categorized into three main types: generation-based, auxiliary property-based, and contrast-based methods \cite{liu2022graph}. Generation-based methods utilize feature or structure reconstruction as the loss function to extract embeddings with strong generalization properties (\textit{e.g.}, GAE \cite{kipf2016variational}, GraphMAE \cite{hou2022graphmae}, and GraphMAE2 \cite{hou2023graphmae2}). Auxiliary property-based methods introduce new attributive or structural properties as supervision signals, such as clustering pseudo labels utilized by M3S \cite{sun2020multi}. Contrast-based methods define positive and negative embedding pairs and aim to bring positive pairs closer while pushing negative pairs apart. 
Among these three categories, contrast-based methods have garnered the most popularity and achieved notable successes \cite{grill2020bootstrap, hassani2020contrastive, sun2019infograph, velivckovic2018deep, you2020graphCL, xia2022simgrace, zhu2022rosa}. However, despite the success of these methods across various graph tasks, none have managed to achieve the objective of pretraining on multiple graph datasets belonging to different domains. Consequently, the efficacy of pretraining remains constrained by the size and diversity of the source data. This remains an open question in the graph pretraining field.

\noindent \textbf{Graph Transfer Learning.} Recent years have witnessed significant advancements in GNNs. However, adapting pre-trained GNNs to diverse downstream tasks efficiently remains a challenge. Traditionally, finetuning, as presented in \cite{lee2017transfer, zhuang2020comprehensive,chen2024graphwiz}, has dominated this field. This approach leverages a pretrained GNN as a foundation, finetuning either the final layer or the entire model for the specific task. Finetuning has consistently achieved state-of-the-art (SOTA) performance.

However, recent exploration has led to the emergence of graph prompts \cite{sun2023all, sun2023graph, wu2023survey, li2023survey} as a compelling alternative, drawing inspiration from the NLP community's `prompting' paradigm \cite{liu2023pre,li2022community,xiong2024TGLLM}. As detailed in \cite{sun2023all}, a graph prompt comprises three key elements: \textit{prompt tokens}, which contain the prompt content in the form of vectors; \textit{token structures}, which depicts how the tokens are connected; \textit{inserting patterns}, which fuses the graph prompt with the target data. By carefully crafting these components, often through learning-based approaches, graph prompts can effectively transfer knowledge from the pretrained model to new tasks. While not guaranteed to outperform finetuning in every scenario, graph prompts excel in terms of efficiency due to their minimal parameter footprint.
This shift towards graph prompts highlights the ongoing exploration of efficient and effective knowledge transfer within the realm of GNNs. Further research is warranted to refine prompt construction strategies and assess their broader applicability across diverse downstream tasks.
\vspace{-1em}
\section{Conclusion}

This study delves into the intricate phenomenon of negative transfer within the field of graph learning, shedding light on its complexities through a rigorous analysis. In response, we introduce an innovative pretraining approach named GCOPE, devised to effectively mitigate negative transfer effects. GCOPE leverages coordinators to seamlessly amalgamate disparate graphs, establishing interconnections and aligning their features. Our thorough experimentation, conducted across a spectrum of homophilic and heterophilic datasets, vividly demonstrates the efficacy of our proposed method.

Despite the notable success achieved by GCOPE, it is prudent to acknowledge the potential limitations stemming from the non-parametric nature of SVD. This aspect may constrain the method's generalizability across diverse datasets. As such, our future research endeavors will be directed towards devising a more robust learning-based feature projection pipeline, which naturally understands and aligns features of graphs from a variety of domains.


\vspace{-0.5em}
\begin{acks}
This Research of Li was supported by NSFC Grant No. 62206067 and Guangzhou-HKUST(GZ) Joint Funding Scheme 2023A03J0673. The research of Cheng was supported in part by project \#MMT-p2-23 of the Shun Hing Institute of Advanced Engineering, The Chinese University of Hong Kong and by grants from the Research Grant Council of the Hong Kong Special Administrative Region, China (No. CUHK 14217622).
\end{acks}
\vspace{-0.5em}

\bibliographystyle{ACM-Reference-Format}
\bibliography{arxiv}

\appendix


\setcounter{section}{0}
\setcounter{figure}{0}
\setcounter{table}{0}
\makeatletter 
\renewcommand{\thefigure}{A\arabic{figure}}
\renewcommand{\theHfigure}{A\arabic{figure}}
\renewcommand{\thetable}{A\arabic{table}}
\renewcommand{\theHtable}{A\arabic{table}}

\begin{table*}[t]
\renewcommand\arraystretch{1.2}
\caption{Cross-domain transfer learning performance (mean±std Acc/AUC/F1) on homophilic datasets (C-way 3-shot). IMP (\%): the average improvement of GCOPE over the rest. GCL and Sim respectively represent GraphCL and SimGRACE.}
\label{tab:rq1 homo 3 shot}
\vspace{-0.5em}
\resizebox{1.0\textwidth}{!}{
\begin{tabular}{cc|ccc|ccc|ccc|ccc|ccc}
\cmidrule[1.2pt]{1-17}
\multirow{2}{*}{\begin{tabular}[c]{@{}c@{}}Training\\ schemes\end{tabular}}                   & \multirow{2}{*}{Methods} & \multicolumn{3}{c|}{Cora} & \multicolumn{3}{c|}{Citeseer} & \multicolumn{3}{c|}{Pubmed} & \multicolumn{3}{c|}{Computers} & \multicolumn{3}{c}{Photos} \\
                                                                                              &                          & Acc     & AUC    & F1     & Acc      & AUC      & F1      & Acc     & AUC     & F1      & Acc      & AUC      & F1       & Acc     & AUC     & F1     \\ \midrule
\multirow{4}{*}{supervised}                                                                   & GCN                       & 0.4493\text{\scriptsize ±.05}                         & 0.7911\text{\scriptsize ±.02}                         & 0.4425\text{\scriptsize ±.04}                         & 0.4336\text{\scriptsize ±.05}                         & 0.7580\text{\scriptsize ±.04}                         & 0.4131\text{\scriptsize ±.04}                         & 0.5565\text{\scriptsize ±.06}                         & 0.7161\text{\scriptsize ±.05}                         & 0.5343\text{\scriptsize ±.06}                         & 0.4516\text{\scriptsize ±.04}                         & 0.7973\text{\scriptsize ±.01}                         & 0.4101\text{\scriptsize ±.02}                         & 0.5816\text{\scriptsize ±.04}                         & 0.8786\text{\scriptsize ±.01}                         & 0.5261\text{\scriptsize ±.05}                         \\
                                                                                               & GAT                       & 0.4741\text{\scriptsize ±.01}                         & 0.8035\text{\scriptsize ±.01}                         & 0.4420\text{\scriptsize ±.01}                         & 0.4813\text{\scriptsize ±.02}                         & 0.7634\text{\scriptsize ±.01}                         & 0.4470\text{\scriptsize ±.02}                         & 0.6250\text{\scriptsize ±.01}                         & 0.7803\text{\scriptsize ±.02}                         & 0.5877\text{\scriptsize ±.04}                         & 0.4077\text{\scriptsize ±.07}                         & 0.7951\text{\scriptsize ±.03}                         & 0.3865\text{\scriptsize ±.02}                         & 0.5935\text{\scriptsize ±.04}                         & 0.8789\text{\scriptsize ±.02}                         & 0.5595\text{\scriptsize ±.05}                         \\
                                                                                               & BWGNN                     & 0.4887\text{\scriptsize ±.01}                         & 0.8137\text{\scriptsize ±.01}                         & 0.4737\text{\scriptsize ±.01}                         & 0.3840\text{\scriptsize ±.05}                         & 0.7176\text{\scriptsize ±.03}                         & 0.3608\text{\scriptsize ±.05}                         & 0.5423\text{\scriptsize ±.03}                         & 0.6940\text{\scriptsize ±.02}                         & 0.5171\text{\scriptsize ±.03}                         & 0.4554\text{\scriptsize ±.04}                         & 0.8022\text{\scriptsize ±.01}                         & 0.4132\text{\scriptsize ±.06}                         & 0.5749\text{\scriptsize ±.03}                         & 0.8670\text{\scriptsize ±.00}                         & 0.5308\text{\scriptsize ±.01}                         \\
                                                                                              & FAGCN                     & 0.6517\text{\scriptsize ±.01} & 0.8926\text{\scriptsize ±.00} & 0.6336\text{\scriptsize ±.01} & 0.6217\text{\scriptsize ±.02} & 0.8416\text{\scriptsize ±.00} & 0.5888\text{\scriptsize ±.02} & 0.5695\text{\scriptsize ±.03} & 0.6267\text{\scriptsize ±.04} & 0.5201\text{\scriptsize ±.02} & 0.5588\text{\scriptsize ±.06} & 0.8695\text{\scriptsize ±.02} & 0.5300\text{\scriptsize ±.04} & 0.6582\text{\scriptsize ±.02} & 0.9123\text{\scriptsize ±.00} & 0.6431\text{\scriptsize ±.01} \\ \midrule
\multirow{4}{*}{\begin{tabular}[c]{@{}c@{}}IP \\ + \\ finetuning\end{tabular}}                & GCL+GCN             & 0.4414\text{\scriptsize ±.06}                         & 0.7540\text{\scriptsize ±.03}                         & 0.4155\text{\scriptsize ±.04}                         & 0.4099\text{\scriptsize ±.04}                         & 0.7270\text{\scriptsize ±.04}                         & 0.3836\text{\scriptsize ±.07}                         & 0.5565\text{\scriptsize ±.03}                         & 0.7374\text{\scriptsize ±.01}                         & 0.5334\text{\scriptsize ±.02}                         & 0.4363\text{\scriptsize ±.03}                         & 0.8222\text{\scriptsize ±.01}                         & 0.4470\text{\scriptsize ±.01}                         & 0.5918\text{\scriptsize ±.04}                         & 0.8949\text{\scriptsize ±.01}                         & 0.5868\text{\scriptsize ±.03}                         \\
                                                                                               & GCL+FAGCN             & 0.6252\text{\scriptsize ±.02}                         & 0.8945\text{\scriptsize ±.00}                         & 0.6129\text{\scriptsize ±.02}                         & 0.6099\text{\scriptsize ±.02}                         & 0.8469\text{\scriptsize ±.00}                         & 0.5767\text{\scriptsize ±.02}                         & 0.5374\text{\scriptsize ±.02}                         & 0.6134\text{\scriptsize ±.02}                         & 0.4643\text{\scriptsize ±.04}                         & 0.5300\text{\scriptsize ±.06}                         & 0.8799\text{\scriptsize ±.00}                         & 0.5226\text{\scriptsize ±.03}                         & 0.6925\text{\scriptsize ±.02}                         & 0.9247\text{\scriptsize ±.00}                         & 0.6575\text{\scriptsize ±.01}                         \\
                                                                                               & Sim+GCN            & 0.4290\text{\scriptsize ±.02}                         & 0.7565\text{\scriptsize ±.02}                         & 0.3911\text{\scriptsize ±.02}                         & 0.3816\text{\scriptsize ±.05}                         & 0.6938\text{\scriptsize ±.04}                         & 0.3375\text{\scriptsize ±.08}                         & 0.5458\text{\scriptsize ±.04}                         & 0.7008\text{\scriptsize ±.02}                         & 0.5083\text{\scriptsize ±.04}                         & 0.4674\text{\scriptsize ±.02}                         & 0.8057\text{\scriptsize ±.02}                         & 0.4061\text{\scriptsize ±.07}                         & 0.5561\text{\scriptsize ±.03}                         & 0.8525\text{\scriptsize ±.01}                         & 0.5131\text{\scriptsize ±.03}                         \\
                                                                                              & Sim+FAGCN            & 0.6758\text{\scriptsize ±.02} & 0.9004\text{\scriptsize ±.01} & 0.6467\text{\scriptsize ±.02} & 0.5333\text{\scriptsize ±.02} & 0.8064\text{\scriptsize ±.00} & 0.4992\text{\scriptsize ±.02} & 0.5696\text{\scriptsize ±.02} & 0.6201\text{\scriptsize ±.03} & 0.5031\text{\scriptsize ±.05} & 0.5164\text{\scriptsize ±.03} & 0.8531\text{\scriptsize ±.01} & 0.5111\text{\scriptsize ±.02} & 0.6532\text{\scriptsize ±.01} & 0.9301\text{\scriptsize ±.00} & 0.6482\text{\scriptsize ±.01} \\ \midrule
\multirow{4}{*}{\begin{tabular}[c]{@{}c@{}}GCOPE\\ +\\ finetuning\end{tabular}}               & GCL+GCN             & 0.4327\text{\scriptsize ±.03}                         & 0.7560\text{\scriptsize ±.04}                         & 0.4131\text{\scriptsize ±.03}                         & 0.4273\text{\scriptsize ±.02}                         & 0.7392\text{\scriptsize ±.02}                         & 0.3871\text{\scriptsize ±.03}                         & 0.5681\text{\scriptsize ±.02}                         & 0.7314\text{\scriptsize ±.01}                         & 0.5103\text{\scriptsize ±.05}                         & 0.4808\text{\scriptsize ±.02}                         & 0.8357\text{\scriptsize ±.01}                         & 0.4424\text{\scriptsize ±.04}                         & 0.6266\text{\scriptsize ±.02}                         & 0.8909\text{\scriptsize ±.01}                         & 0.5754\text{\scriptsize ±.03}                         \\
                                                                                               & GCL+FAGCN             & 0.6857\text{\scriptsize ±.02}                         & 0.9166\text{\scriptsize ±.00}                         & 0.6721\text{\scriptsize ±.02}                         & 0.6148\text{\scriptsize ±.02}                         & 0.8431\text{\scriptsize ±.01}                         & 0.5752\text{\scriptsize ±.02}                         & 0.5490\text{\scriptsize ±.02}                         & 0.6636\text{\scriptsize ±.01}                         & 0.4878\text{\scriptsize ±.06}                         & 0.5493\text{\scriptsize ±.03}                         & 0.8682\text{\scriptsize ±.01}                         & 0.5269\text{\scriptsize ±.05}                         & 0.6873\text{\scriptsize ±.03}                         & 0.9172\text{\scriptsize ±.01}                         & 0.6556\text{\scriptsize ±.01}                         \\
                                                                                               & Sim+GCN            & 0.4187\text{\scriptsize ±.01}                         & 0.7648\text{\scriptsize ±.03}                         & 0.3841\text{\scriptsize ±.05}                         & 0.4301\text{\scriptsize ±.00}                         & 0.7212\text{\scriptsize ±.00}                         & 0.3952\text{\scriptsize ±.03}                         & 0.5606\text{\scriptsize ±.05}                         & 0.7418\text{\scriptsize ±.04}                         & 0.5246\text{\scriptsize ±.06}                         & 0.4432\text{\scriptsize ±.03}                         & 0.7990\text{\scriptsize ±.01}                         & 0.3968\text{\scriptsize ±.04}                         & 0.5890\text{\scriptsize ±.01}                         & 0.8841\text{\scriptsize ±.01}                         & 0.5441\text{\scriptsize ±.03}                         \\
                                                                                              & Sim+FAGCN            & 0.6798\text{\scriptsize ±.04}                         & 0.9010\text{\scriptsize ±.02}                         & 0.6617\text{\scriptsize ±.04}                         & 0.6469\text{\scriptsize ±.02}                         & 0.8602\text{\scriptsize ±.00}                         & 0.6120\text{\scriptsize ±.02}                         & 0.6048\text{\scriptsize ±.03}                         & 0.7081\text{\scriptsize ±.07}                         & 0.5700\text{\scriptsize ±.05}                         & 0.5742\text{\scriptsize ±.02}                         & 0.8684\text{\scriptsize ±.01}                         & 0.5480\text{\scriptsize ±.00}                         & 0.6938\text{\scriptsize ±.01}                         & 0.9343\text{\scriptsize ±.00}                         & 0.6683\text{\scriptsize ±.00}                         \\ \midrule
\multicolumn{2}{c|}{IMP (\%)}                                                                                             & 4.69\%                                    & 1.07\%                                    & 5.03\%                                    & 9.93\%                                    & 2.81\%                                    & 9.21\%                                    & 1.39\%                                    & 3.66\%                                    & 0.41\%                                    & 7.10\%                                    & 1.78\%                                    & 5.56\%                                    & 5.95\%                                    & 1.60\%                                    & 5.18\%                                    \\
\cmidrule[1.2pt]{1-17}
\end{tabular}
}
\vspace{-0.5em}
\end{table*}

\begin{table*}[t]
\renewcommand\arraystretch{1.2}
\caption{Cross-domain transfer learning performance (mean±std Acc/AUC/F1) on heterophilic datasets (C-way-3-shot). IMP (\%): the average improvement of GCOPE over the rest. GCL and Sim respectively represent GraphCL and SimGRACE.}
\label{tab:rq1 hetero 3 shot}
\vspace{-0.5em}
\resizebox{1.0\textwidth}{!}{
\begin{tabular}{cc|ccc|ccc|ccc|ccc|ccc}
\cmidrule[1.2pt]{1-17}
\multirow{2}{*}{\begin{tabular}[c]{@{}c@{}}Training\\ schemes\end{tabular}}                   & \multirow{2}{*}{Methods} & \multicolumn{3}{c|}{Wisconsin} & \multicolumn{3}{c|}{Texas} & \multicolumn{3}{c|}{Cornell} & \multicolumn{3}{c|}{Chameleon} & \multicolumn{3}{c}{Squirrel} \\
                                                                                              &                          & Acc      & AUC      & F1       & Acc     & AUC     & F1     & Acc      & AUC     & F1      & Acc      & AUC      & F1       & Acc      & AUC      & F1      \\ \midrule
\multirow{4}{*}{supervised}                                                                   & GCN                       & 0.6642\text{\scriptsize ±.05}                         & 0.8406\text{\scriptsize ±.02}                         & 0.6136\text{\scriptsize ±.05}                         & 0.7085\text{\scriptsize ±.03}                         & 0.6826\text{\scriptsize ±.01}                         & 0.5778\text{\scriptsize ±.08}                         & 0.5444\text{\scriptsize ±.12}                         & 0.8084\text{\scriptsize ±.05}                         & 0.5276\text{\scriptsize ±.06}                         & 0.2340\text{\scriptsize ±.02}                         & 0.5359\text{\scriptsize ±.03}                         & 0.1566\text{\scriptsize ±.05}                         & 0.2161\text{\scriptsize ±.00}                         & 0.5228\text{\scriptsize ±.00}                         & 0.1916\text{\scriptsize ±.01}                         \\
                                                                                               & GAT                       & 0.5783\text{\scriptsize ±.05}                         & 0.7976\text{\scriptsize ±.02}                         & 0.4903\text{\scriptsize ±.05}                         & 0.6732\text{\scriptsize ±.04}                         & 0.6320\text{\scriptsize ±.03}                         & 0.5533\text{\scriptsize ±.02}                         & 0.4411\text{\scriptsize ±.11}                         & 0.7879\text{\scriptsize ±.06}                         & 0.4410\text{\scriptsize ±.07}                         & 0.2400\text{\scriptsize ±.01}                         & 0.5401\text{\scriptsize ±.01}                         & 0.1782\text{\scriptsize ±.04}                         & 0.2011\text{\scriptsize ±.01}                         & 0.5064\text{\scriptsize ±.01}                         & 0.1536\text{\scriptsize ±.03}                         \\
                                                                                               & BWGNN                     & 0.5868\text{\scriptsize ±.05}                         & 0.8212\text{\scriptsize ±.06}                         & 0.4435\text{\scriptsize ±.09}                         & 0.6654\text{\scriptsize ±.09}                         & 0.6694\text{\scriptsize ±.05}                         & 0.4401\text{\scriptsize ±.24}                         & 0.5417\text{\scriptsize ±.05}                         & 0.7871\text{\scriptsize ±.07}                         & 0.4799\text{\scriptsize ±.05}                         & 0.2245\text{\scriptsize ±.04}                         & 0.4953\text{\scriptsize ±.02}                         & 0.1816\text{\scriptsize ±.02}                         & 0.2096\text{\scriptsize ±.00}                         & 0.5075\text{\scriptsize ±.01}                         & 0.1635\text{\scriptsize ±.02}                         \\
                                                                                              & FAGCN                     & 0.7104\text{\scriptsize ±.01} & 0.8655\text{\scriptsize ±.01} & 0.5972\text{\scriptsize ±.02} & 0.7163\text{\scriptsize ±.04} & 0.6728\text{\scriptsize ±.01} & 0.6105\text{\scriptsize ±.04} & 0.5722\text{\scriptsize ±.10} & 0.8638\text{\scriptsize ±.02} & 0.5835\text{\scriptsize ±.09} & 0.2564\text{\scriptsize ±.02} & 0.5425\text{\scriptsize ±.02} & 0.2042\text{\scriptsize ±.02} & 0.2053\text{\scriptsize ±.01} & 0.5064\text{\scriptsize ±.01} & 0.1738\text{\scriptsize ±.01} \\ \midrule
\multirow{4}{*}{\begin{tabular}[c]{@{}c@{}}IP \\ +\\ finetuning\end{tabular}}                 & GCL+GCN             & 0.5840\text{\scriptsize ±.07}                         & 0.8030\text{\scriptsize ±.08}                         & 0.4231\text{\scriptsize ±.19}                         & 0.6510\text{\scriptsize ±.08}                         & 0.6508\text{\scriptsize ±.02}                         & 0.5709\text{\scriptsize ±.13}                         & 0.5470\text{\scriptsize ±.09}                         & 0.8111\text{\scriptsize ±.03}                         & 0.4722\text{\scriptsize ±.08}                         & 0.2245\text{\scriptsize ±.02}                         & 0.5092\text{\scriptsize ±.02}                         & 0.1635\text{\scriptsize ±.03}                         & 0.2070\text{\scriptsize ±.01}                         & 0.5027\text{\scriptsize ±.00}                         & 0.1551\text{\scriptsize ±.04}                         \\
                                                                                               & GCL+FAGCN             & 0.6547\text{\scriptsize ±.04}                         & 0.8624\text{\scriptsize ±.01}                         & 0.5667\text{\scriptsize ±.04}                         & 0.7242\text{\scriptsize ±.03}                         & 0.6668\text{\scriptsize ±.00}                         & 0.6434\text{\scriptsize ±.07}                         & 0.5881\text{\scriptsize ±.06}                         & 0.8756\text{\scriptsize ±.01}                         & 0.6183\text{\scriptsize ±.06}                         & 0.2590\text{\scriptsize ±.02}                         & 0.5439\text{\scriptsize ±.01}                         & 0.2180\text{\scriptsize ±.01}                         & 0.2039\text{\scriptsize ±.01}                         & 0.5077\text{\scriptsize ±.00}                         & 0.1558\text{\scriptsize ±.04}                         \\
                                                                                               & Sim+GCN            & 0.5311\text{\scriptsize ±.08}                         & 0.8151\text{\scriptsize ±.04}                         & 0.4518\text{\scriptsize ±.09}                         & 0.6797\text{\scriptsize ±.05}                         & 0.6773\text{\scriptsize ±.03}                         & 0.4847\text{\scriptsize ±.06}                         & 0.4848\text{\scriptsize ±.09}                         & 0.8138\text{\scriptsize ±.06}                         & 0.4507\text{\scriptsize ±.05}                         & 0.2252\text{\scriptsize ±.02}                         & 0.5140\text{\scriptsize ±.01}                         & 0.1871\text{\scriptsize ±.02}                         & 0.2062\text{\scriptsize ±.01}                         & 0.5091\text{\scriptsize ±.00}                         & 0.1553\text{\scriptsize ±.04}                         \\
                                                                                              & Sim+FAGCN            & 0.6349\text{\scriptsize ±.01} & 0.8876\text{\scriptsize ±.00} & 0.5512\text{\scriptsize ±.01} & 0.7464\text{\scriptsize ±.01} & 0.6750\text{\scriptsize ±.00} & 0.6949\text{\scriptsize ±.01} & 0.5709\text{\scriptsize ±.09} & 0.8829\text{\scriptsize ±.01} & 0.5953\text{\scriptsize ±.07} & 0.2376\text{\scriptsize ±.02} & 0.5251\text{\scriptsize ±.02} & 0.1921\text{\scriptsize ±.02} & 0.2027\text{\scriptsize ±.00} & 0.5109\text{\scriptsize ±.01} & 0.1587\text{\scriptsize ±.05} \\ \midrule
\multirow{4}{*}{\begin{tabular}[c]{@{}c@{}}GCOPE\\ +\\ finetuning\end{tabular}}               & GCL+GCN             & 0.5943\text{\scriptsize ±.05}                         & 0.8351\text{\scriptsize ±.03}                         & 0.4876\text{\scriptsize ±.06}                         & 0.7490\text{\scriptsize ±.05}                         & 0.7029\text{\scriptsize ±.02}                         & 0.5909\text{\scriptsize ±.11}                         & 0.5351\text{\scriptsize ±.05}                         & 0.8307\text{\scriptsize ±.04}                         & 0.4823\text{\scriptsize ±.07}                         & 0.2276\text{\scriptsize ±.01}                         & 0.5354\text{\scriptsize ±.01}                         & 0.1573\text{\scriptsize ±.02}                         & 0.2110\text{\scriptsize ±.00}                         & 0.5189\text{\scriptsize ±.01}                         & 0.1653\text{\scriptsize ±.02}                         \\
                                                                                               & GCL+FAGCN             & 0.6481\text{\scriptsize ±.03}                         & 0.8545\text{\scriptsize ±.00}                         & 0.5634\text{\scriptsize ±.02}                         & 0.7033\text{\scriptsize ±.03}                         & 0.6554\text{\scriptsize ±.01}                         & 0.5997\text{\scriptsize ±.04}                         & 0.5669\text{\scriptsize ±.04}                         & 0.8566\text{\scriptsize ±.01}                         & 0.6024\text{\scriptsize ±.03}                         & 0.2442\text{\scriptsize ±.01}                         & 0.5454\text{\scriptsize ±.01}                         & 0.2105\text{\scriptsize ±.00}                         & 0.1985\text{\scriptsize ±.00}                         & 0.5039\text{\scriptsize ±.00}                         & 0.1456\text{\scriptsize ±.03}                         \\
                                                                                               & Sim+GCN            & 0.5726\text{\scriptsize ±.07}                         & 0.8299\text{\scriptsize ±.05}                         & 0.4283\text{\scriptsize ±.17}                         & 0.7882\text{\scriptsize ±.04}                         & 0.7060\text{\scriptsize ±.01}                         & 0.6635\text{\scriptsize ±.06}                         & 0.6159\text{\scriptsize ±.04}                         & 0.8656\text{\scriptsize ±.02}                         & 0.5810\text{\scriptsize ±.04}                         & 0.2084\text{\scriptsize ±.01}                         & 0.5039\text{\scriptsize ±.01}                         & 0.1660\text{\scriptsize ±.03}                         & 0.2159\text{\scriptsize ±.01}                         & 0.5135\text{\scriptsize ±.00}                         & 0.1746\text{\scriptsize ±.02}                         \\
                                                                                              & Sim+FAGCN            & 0.7575\text{\scriptsize ±.01}                         & 0.9279\text{\scriptsize ±.00}                         & 0.6998\text{\scriptsize ±.01}                         & 0.8105\text{\scriptsize ±.01}                         & 0.7140\text{\scriptsize ±.01}                         & 0.7604\text{\scriptsize ±.01}                         & 0.7868\text{\scriptsize ±.02}                         & 0.9237\text{\scriptsize ±.00}                         & 0.7490\text{\scriptsize ±.03}                         & 0.2362\text{\scriptsize ±.02}                         & 0.5345\text{\scriptsize ±.01}                         & 0.2052\text{\scriptsize ±.01}                         & 0.2037\text{\scriptsize ±.00}                         & 0.5076\text{\scriptsize ±.00}                         & 0.1507\text{\scriptsize ±.04}                         \\ \midrule  
\multicolumn{2}{c|}{IMP(\%)}    & 4.06\%                                    & 3.02\%                                          & 5.34\%                                    & 9.66\%                                    & 4.32\%                                    & 14.28\%                                    & 16.76\%                                    & 4.87\%                                    & 15.85\%                                    & -3.60\%                                    & 0.68\%                                    & -0.22\%                                    & 0.38\%                                    & 0.35\%                                    & -2.68\%                                          \\
\cmidrule[1.2pt]{1-17}
\end{tabular}
}
\vspace{-0.5em}
\end{table*}

\begin{table*}[t]
\renewcommand\arraystretch{1.2}
\caption{Cross-domain transfer learning performance (mean±std Acc/AUC/F1) on homophilic datasets (C-way-5-shot). IMP (\%): the average improvement of GCOPE over the rest. GCL and Sim respectively represent GraphCL and SimGRACE.}
\label{tab:rq1 homo 5 shot}
\vspace{-0.5em}
\resizebox{1.0\textwidth}{!}{
\begin{tabular}{cc|ccc|ccc|ccc|ccc|ccc}
\cmidrule[1.2pt]{1-17}
\multirow{2}{*}{\begin{tabular}[c]{@{}c@{}}Training\\ schemes\end{tabular}}                   & \multirow{2}{*}{Methods} & \multicolumn{3}{c|}{Cora} & \multicolumn{3}{c|}{Citeseer} & \multicolumn{3}{c|}{Pubmed} & \multicolumn{3}{c|}{Computers} & \multicolumn{3}{c}{Photos} \\
                                                                                              &                          & Acc     & AUC    & F1     & Acc      & AUC      & F1      & Acc     & AUC     & F1      & Acc      & AUC      & F1       & Acc     & AUC     & F1     \\ \midrule
\multirow{4}{*}{supervised}                                                                   & GCN                       & 0.6105\text{\scriptsize ±.03}                         & 0.8800\text{\scriptsize ±.03}                         & 0.6078\text{\scriptsize ±.04}                         & 0.4554\text{\scriptsize ±.04}                         & 0.7478\text{\scriptsize ±.04}                         & 0.4361\text{\scriptsize ±.02}                         & 0.6135\text{\scriptsize ±.02}                         & 0.7644\text{\scriptsize ±.03}                         & 0.5786\text{\scriptsize ±.05}                         & 0.4274\text{\scriptsize ±.05}                         & 0.8126\text{\scriptsize ±.03}                         & 0.4313\text{\scriptsize ±.03}                         & 0.6041\text{\scriptsize ±.04}                         & 0.8931\text{\scriptsize ±.00}                         & 0.5817\text{\scriptsize ±.03}                         \\
                                                                                               & GAT                       & 0.5979\text{\scriptsize ±.03}                         & 0.8652\text{\scriptsize ±.02}                         & 0.5727\text{\scriptsize ±.05}                         & 0.5048\text{\scriptsize ±.02}                         & 0.7927\text{\scriptsize ±.01}                         & 0.4837\text{\scriptsize ±.03}                         & 0.6165\text{\scriptsize ±.01}                         & 0.7537\text{\scriptsize ±.02}                         & 0.5772\text{\scriptsize ±.03}                         & 0.5108\text{\scriptsize ±.04}                         & 0.8423\text{\scriptsize ±.01}                         & 0.4666\text{\scriptsize ±.04}                         & 0.6257\text{\scriptsize ±.02}                         & 0.9022\text{\scriptsize ±.00}                         & 0.5936\text{\scriptsize ±.02}                         \\
                                                                                               & BWGNN                     & 0.6502\text{\scriptsize ±.03}                         & 0.8869\text{\scriptsize ±.01}                         & 0.6378\text{\scriptsize ±.02}                         & 0.5228\text{\scriptsize ±.03}                         & 0.8096\text{\scriptsize ±.02}                         & 0.4886\text{\scriptsize ±.04}                         & 0.5839\text{\scriptsize ±.02}                         & 0.7274\text{\scriptsize ±.04}                         & 0.5284\text{\scriptsize ±.05}                         & 0.5390\text{\scriptsize ±.03}                         & 0.8483\text{\scriptsize ±.02}                         & 0.4919\text{\scriptsize ±.04}                         & 0.6157\text{\scriptsize ±.03}                         & 0.8891\text{\scriptsize ±.00}                         & 0.5839\text{\scriptsize ±.03}                         \\
                                                                                              & FAGCN                     & 0.7183\text{\scriptsize ±.01} & 0.9139\text{\scriptsize ±.01} & 0.7144\text{\scriptsize ±.01} & 0.6721\text{\scriptsize ±.00} & 0.8597\text{\scriptsize ±.00} & 0.6390\text{\scriptsize ±.00} & 0.5965\text{\scriptsize ±.01} & 0.7445\text{\scriptsize ±.03} & 0.5719\text{\scriptsize ±.03} & 0.6054\text{\scriptsize ±.01} & 0.8886\text{\scriptsize ±.01} & 0.5851\text{\scriptsize ±.02} & 0.6655\text{\scriptsize ±.02} & 0.9107\text{\scriptsize ±.00} & 0.6435\text{\scriptsize ±.01} \\ \midrule
\multirow{4}{*}{\begin{tabular}[c]{@{}c@{}}IP \\ + \\ finetuning\end{tabular}}                & GCL+GCN             & 0.6351\text{\scriptsize ±.01}                         & 0.8807\text{\scriptsize ±.00}                         & 0.6159\text{\scriptsize ±.01}                         & 0.5347\text{\scriptsize ±.03}                         & 0.8068\text{\scriptsize ±.01}                         & 0.5203\text{\scriptsize ±.03}                         & 0.5719\text{\scriptsize ±.03}                         & 0.6934\text{\scriptsize ±.04}                         & 0.5382\text{\scriptsize ±.03}                         & 0.4849\text{\scriptsize ±.07}                         & 0.8576\text{\scriptsize ±.01}                         & 0.4760\text{\scriptsize ±.07}                         & 0.6183\text{\scriptsize ±.02}                         & 0.8993\text{\scriptsize ±.00}                         & 0.6039\text{\scriptsize ±.00}                         \\
                                                                                               & GCL+FAGCN             & 0.6948\text{\scriptsize ±.02} & 0.9059\text{\scriptsize ±.01} & 0.6893\text{\scriptsize ±.03} & 0.6768\text{\scriptsize ±.01} & 0.8690\text{\scriptsize ±.00} & 0.6530\text{\scriptsize ±.01} & 0.6024\text{\scriptsize ±.01} & 0.7596\text{\scriptsize ±.02} & 0.5876\text{\scriptsize ±.01} & 0.6217\text{\scriptsize ±.01} & 0.8806\text{\scriptsize ±.00} & 0.5868\text{\scriptsize ±.00} & 0.6991\text{\scriptsize ±.01} & 0.9083\text{\scriptsize ±.00} & 0.6509\text{\scriptsize ±.00} \\
                                                                                               & Sim+GCN            & 0.5596\text{\scriptsize ±.04}                         & 0.8520\text{\scriptsize ±.01}                         & 0.5441\text{\scriptsize ±.03}                         & 0.5180\text{\scriptsize ±.02}                         & 0.7962\text{\scriptsize ±.00}                         & 0.4905\text{\scriptsize ±.03}                         & 0.5064\text{\scriptsize ±.02}                         & 0.6503\text{\scriptsize ±.03}                         & 0.4582\text{\scriptsize ±.06}                         & 0.4817\text{\scriptsize ±.02}                         & 0.8464\text{\scriptsize ±.01}                         & 0.4525\text{\scriptsize ±.02}                         & 0.6183\text{\scriptsize ±.03}                         & 0.9102\text{\scriptsize ±.00}                         & 0.5964\text{\scriptsize ±.03}                         \\
                                                                                              & Sim+FAGCN            & 0.7247\text{\scriptsize ±.01} & 0.9183\text{\scriptsize ±.00} & 0.7179\text{\scriptsize ±.02} & 0.6584\text{\scriptsize ±.01} & 0.8693\text{\scriptsize ±.00} & 0.6384\text{\scriptsize ±.01} & 0.6038\text{\scriptsize ±.01} & 0.7467\text{\scriptsize ±.04} & 0.5690\text{\scriptsize ±.04} & 0.5888\text{\scriptsize ±.01} & 0.8749\text{\scriptsize ±.02} & 0.5505\text{\scriptsize ±.02} & 0.6845\text{\scriptsize ±.02} & 0.9109\text{\scriptsize ±.00} & 0.6380\text{\scriptsize ±.01} \\ \midrule
\multirow{4}{*}{\begin{tabular}[c]{@{}c@{}}GCOPE\\ +\\ finetuning\end{tabular}}               & GCL+GCN             & 0.6079\text{\scriptsize ±.02}                         & 0.8705\text{\scriptsize ±.02}                         & 0.5971\text{\scriptsize ±.03}                         & 0.5745\text{\scriptsize ±.06}                         & 0.8354\text{\scriptsize ±.02}                         & 0.5500\text{\scriptsize ±.06}                         & 0.5883\text{\scriptsize ±.03}                         & 0.7352\text{\scriptsize ±.03}                         & 0.5317\text{\scriptsize ±.05}                         & 0.5552\text{\scriptsize ±.03}                         & 0.8621\text{\scriptsize ±.01}                         & 0.4795\text{\scriptsize ±.06}                         & 0.6494\text{\scriptsize ±.01}                         & 0.9120\text{\scriptsize ±.00}                         & 0.6153\text{\scriptsize ±.03}                         \\
                                                                                               & GCL+FAGCN             & 0.7334\text{\scriptsize ±.01}                         & 0.9247\text{\scriptsize ±.00}                         & 0.7238\text{\scriptsize ±.01}                         & 0.6863\text{\scriptsize ±.02}                         & 0.8778\text{\scriptsize ±.00}                         & 0.6539\text{\scriptsize ±.02}                         & 0.6110\text{\scriptsize ±.00}                         & 0.7552\text{\scriptsize ±.02}                         & 0.5806\text{\scriptsize ±.01}                         & 0.6172\text{\scriptsize ±.01}                         & 0.8950\text{\scriptsize ±.00}                         & 0.5969\text{\scriptsize ±.01}                         & 0.6868\text{\scriptsize ±.00}                         & 0.9206\text{\scriptsize ±.00}                         & 0.6591\text{\scriptsize ±.00}                         \\
                                                                                               & Sim+GCN            & 0.5501\text{\scriptsize ±.02}                         & 0.8522\text{\scriptsize ±.01}                         & 0.5298\text{\scriptsize ±.02}                         & 0.4869\text{\scriptsize ±.01}                         & 0.7884\text{\scriptsize ±.01}                         & 0.4731\text{\scriptsize ±.02}                         & 0.6135\text{\scriptsize ±.04}                         & 0.7684\text{\scriptsize ±.04}                         & 0.5685\text{\scriptsize ±.06}                         & 0.5264\text{\scriptsize ±.00}                         & 0.8588\text{\scriptsize ±.01}                         & 0.4928\text{\scriptsize ±.03}                         & 0.6237\text{\scriptsize ±.03}                         & 0.9105\text{\scriptsize ±.00}                         & 0.5759\text{\scriptsize ±.04}                         \\
                                                                                              & Sim+FAGCN            & 0.7313\text{\scriptsize ±.01}                         & 0.9100\text{\scriptsize ±.02}                         & 0.7189\text{\scriptsize ±.01}                         & 0.6931\text{\scriptsize ±.01}                         & 0.8829\text{\scriptsize ±.00}                         & 0.6631\text{\scriptsize ±.01}                         & 0.5754\text{\scriptsize ±.03}                         & 0.7483\text{\scriptsize ±.03}                         & 0.5541\text{\scriptsize ±.03}                         & 0.6212\text{\scriptsize ±.02}                         & 0.8873\text{\scriptsize ±.01}                         & 0.5710\text{\scriptsize ±.02}                         & 0.6995\text{\scriptsize ±.02}                         & 0.9190\text{\scriptsize ±.00}                         & 0.6544\text{\scriptsize ±.01}                         \\ \midrule
\multicolumn{2}{c|}{IMP (\%)}                                                                                             & 1.05\%                                    & 0.17\%                                    & 0.77\%                                    & 7.45\%                                    & 3.33\%                                    & 7.60\%                                    & 1.74\%                                    & 2.98\%                                    & 1.38\%                                    & 8.93\%                                    & 2.26\%                                    & 5.93\%                                    & 3.66\%                                    & 1.39\%                                    & 2.40\%                                    \\
\cmidrule[1.2pt]{1-17}
\end{tabular}
}
\vspace{-0.5em}
\end{table*}

\begin{table*}[t]
\renewcommand\arraystretch{1.2}
\caption{Cross-domain transfer learning performance (mean±std Acc/AUC/F1) on heterophilic datasets (C-way-5-shot). IMP (\%): the average improvement of GCOPE over the rest. GCL and Sim respectively represent GraphCL and SimGRACE.}
\label{tab:rq1 hetero 5 shot}
\resizebox{1.0\textwidth}{!}{
\begin{tabular}{cc|ccc|ccc|ccc|ccc|ccc}
\cmidrule[1.2pt]{1-17}
\multirow{2}{*}{\begin{tabular}[c]{@{}c@{}}Training\\ schemes\end{tabular}}                   & \multirow{2}{*}{Methods} & \multicolumn{3}{c|}{Wisconsin} & \multicolumn{3}{c|}{Texas} & \multicolumn{3}{c|}{Cornell} & \multicolumn{3}{c|}{Chameleon} & \multicolumn{3}{c}{Squirrel} \\
                                                                                              &                          & Acc      & AUC      & F1       & Acc     & AUC     & F1     & Acc      & AUC     & F1      & Acc      & AUC      & F1       & Acc      & AUC      & F1      \\ \midrule
\multirow{4}{*}{supervised}                                                                   & GCN                       & 0.6374\text{\scriptsize ±.08}                         & 0.8774\text{\scriptsize ±.02}                         & 0.6540\text{\scriptsize ±.06}                         & 0.7117\text{\scriptsize ±.04}                         & 0.6994\text{\scriptsize ±.01}                         & 0.6534\text{\scriptsize ±.04}                         & 0.6239\text{\scriptsize ±.09}                         & 0.8606\text{\scriptsize ±.05}                         & 0.5903\text{\scriptsize ±.07}                         & 0.2577\text{\scriptsize ±.01}                         & 0.5484\text{\scriptsize ±.01}                         & 0.2249\text{\scriptsize ±.01}                         & 0.2209\text{\scriptsize ±.00}                         & 0.5228\text{\scriptsize ±.00}                         & 0.1820\text{\scriptsize ±.02}                         \\
                                                                                               & GAT                       & 0.6128\text{\scriptsize ±.02}                         & 0.8788\text{\scriptsize ±.00}                         & 0.5704\text{\scriptsize ±.05}                         & 0.7517\text{\scriptsize ±.01}                         & 0.6977\text{\scriptsize ±.00}                         & 0.6290\text{\scriptsize ±.06}                         & 0.5549\text{\scriptsize ±.02}                         & 0.8655\text{\scriptsize ±.01}                         & 0.5745\text{\scriptsize ±.05}                         & 0.2147\text{\scriptsize ±.01}                         & 0.5206\text{\scriptsize ±.01}                         & 0.2127\text{\scriptsize ±.01}                         & 0.2083\text{\scriptsize ±.00}                         & 0.5112\text{\scriptsize ±.00}                         & 0.1596\text{\scriptsize ±.03}                         \\
                                                                                               & BWGNN                     & 0.5951\text{\scriptsize ±.00}                         & 0.8745\text{\scriptsize ±.01}                         & 0.5902\text{\scriptsize ±.03}                         & 0.7628\text{\scriptsize ±.04}                         & 0.7236\text{\scriptsize ±.01}                         & 0.6395\text{\scriptsize ±.11}                         & 0.6901\text{\scriptsize ±.07}                         & 0.8963\text{\scriptsize ±.02}                         & 0.6151\text{\scriptsize ±.04}                         & 0.2338\text{\scriptsize ±.01}                         & 0.5386\text{\scriptsize ±.00}                         & 0.1881\text{\scriptsize ±.03}                         & 0.2210\text{\scriptsize ±.01}                         & 0.5319\text{\scriptsize ±.01}                         & 0.1779\text{\scriptsize ±.03}                         \\
                                                                                              & FAGCN                     & 0.5261\text{\scriptsize ±.03} & 0.8678\text{\scriptsize ±.02} & 0.5074\text{\scriptsize ±.03} & 0.7269\text{\scriptsize ±.02} & 0.6945\text{\scriptsize ±.01} & 0.6734\text{\scriptsize ±.02} & 0.5183\text{\scriptsize ±.02} & 0.8841\text{\scriptsize ±.01} & 0.5628\text{\scriptsize ±.03} & 0.2319\text{\scriptsize ±.00} & 0.5212\text{\scriptsize ±.00} & 0.2137\text{\scriptsize ±.00} & 0.2131\text{\scriptsize ±.00} & 0.5162\text{\scriptsize ±.01} & 0.1863\text{\scriptsize ±.01} \\ \midrule
\multirow{4}{*}{\begin{tabular}[c]{@{}c@{}}IP \\ +\\ finetuning\end{tabular}}                 & GCL+GCN             & 0.5941\text{\scriptsize ±.13}                         & 0.9031\text{\scriptsize ±.01}                         & 0.5893\text{\scriptsize ±.14}                         & 0.6331\text{\scriptsize ±.14}                         & 0.6858\text{\scriptsize ±.03}                         & 0.4816\text{\scriptsize ±.14}                         & 0.6141\text{\scriptsize ±.09}                         & 0.8478\text{\scriptsize ±.04}                         & 0.5050\text{\scriptsize ±.05}                         & 0.2500\text{\scriptsize ±.01}                         & 0.5253\text{\scriptsize ±.01}                         & 0.1685\text{\scriptsize ±.01}                         & 0.2144\text{\scriptsize ±.00}                         & 0.5181\text{\scriptsize ±.00}                         & 0.1821\text{\scriptsize ±.02}                         \\
                                                                                               & GCL+FAGCN             & 0.6039\text{\scriptsize ±.05} & 0.9077\text{\scriptsize ±.01} & 0.5963\text{\scriptsize ±.05} & 0.7338\text{\scriptsize ±.00} & 0.7182\text{\scriptsize ±.01} & 0.6045\text{\scriptsize ±.07} & 0.6930\text{\scriptsize ±.03} & 0.9253\text{\scriptsize ±.00} & 0.6839\text{\scriptsize ±.04} & 0.2319\text{\scriptsize ±.00} & 0.5327\text{\scriptsize ±.00} & 0.2196\text{\scriptsize ±.01} & 0.2127\text{\scriptsize ±.00} & 0.5172\text{\scriptsize ±.00} & 0.1931\text{\scriptsize ±.01} \\
                                                                                               & Sim+GCN            & 0.5655\text{\scriptsize ±.05}                         & 0.8860\text{\scriptsize ±.02}                         & 0.5922\text{\scriptsize ±.05}                         & 0.7545\text{\scriptsize ±.01}                         & 0.7312\text{\scriptsize ±.01}                         & 0.6268\text{\scriptsize ±.05}                         & 0.6732\text{\scriptsize ±.04}                         & 0.8717\text{\scriptsize ±.03}                         & 0.5895\text{\scriptsize ±.07}                         & 0.2419\text{\scriptsize ±.01}                         & 0.5354\text{\scriptsize ±.00}                         & 0.1832\text{\scriptsize ±.03}                         & 0.2240\text{\scriptsize ±.00}                         & 0.5321\text{\scriptsize ±.01}                         & 0.1890\text{\scriptsize ±.01}                         \\
                                                                                              & Sim+FAGCN            & 0.6739\text{\scriptsize ±.02} & 0.9166\text{\scriptsize ±.00} & 0.6258\text{\scriptsize ±.04} & 0.7352\text{\scriptsize ±.01} & 0.7162\text{\scriptsize ±.00} & 0.6624\text{\scriptsize ±.06} & 0.7634\text{\scriptsize ±.02} & 0.9572\text{\scriptsize ±.00} & 0.7583\text{\scriptsize ±.02} & 0.2493\text{\scriptsize ±.01} & 0.5400\text{\scriptsize ±.00} & 0.2261\text{\scriptsize ±.00} & 0.2131\text{\scriptsize ±.00} & 0.5135\text{\scriptsize ±.00} & 0.1764\text{\scriptsize ±.01} \\  \midrule
\multirow{4}{*}{\begin{tabular}[c]{@{}c@{}}GCOPE\\ +\\ finetuning\end{tabular}}               & GCL+GCN             & 0.5793\text{\scriptsize ±.08}                         & 0.8864\text{\scriptsize ±.02}                         & 0.5681\text{\scriptsize ±.11}                         & 0.7531\text{\scriptsize ±.04}                         & 0.7315\text{\scriptsize ±.00}                         & 0.6480\text{\scriptsize ±.08}                         & 0.6859\text{\scriptsize ±.15}                         & 0.9129\text{\scriptsize ±.03}                         & 0.6058\text{\scriptsize ±.13}                         & 0.2304\text{\scriptsize ±.01}                         & 0.5264\text{\scriptsize ±.00}                         & 0.2126\text{\scriptsize ±.01}                         & 0.2228\text{\scriptsize ±.01}                         & 0.5319\text{\scriptsize ±.00}                         & 0.1705\text{\scriptsize ±.04}                         \\
                                                                                               & GCL+FAGCN             & 0.6079\text{\scriptsize ±.04}                         & 0.8971\text{\scriptsize ±.00}                         & 0.5870\text{\scriptsize ±.04}                         & 0.7393\text{\scriptsize ±.01}                         & 0.6989\text{\scriptsize ±.01}                         & 0.6736\text{\scriptsize ±.02}                         & 0.6352\text{\scriptsize ±.03}                         & 0.9161\text{\scriptsize ±.00}                         & 0.6302\text{\scriptsize ±.03}                         & 0.2350\text{\scriptsize ±.00}                         & 0.5337\text{\scriptsize ±.00}                         & 0.2102\text{\scriptsize ±.01}                         & 0.2134\text{\scriptsize ±.00}                         & 0.5220\text{\scriptsize ±.00}                         & 0.1943\text{\scriptsize ±.00}                         \\
                                                                                               & Sim+GCN            & 0.6099\text{\scriptsize ±.06}                         & 0.8923\text{\scriptsize ±.01}                         & 0.5884\text{\scriptsize ±.09}                         & 0.7407\text{\scriptsize ±.04}                         & 0.7226\text{\scriptsize ±.01}                         & 0.6326\text{\scriptsize ±.13}                         & 0.5563\text{\scriptsize ±.05}                         & 0.8352\text{\scriptsize ±.05}                         & 0.5248\text{\scriptsize ±.07}                         & 0.2486\text{\scriptsize ±.01}                         & 0.5458\text{\scriptsize ±.01}                         & 0.1860\text{\scriptsize ±.03}                         & 0.2208\text{\scriptsize ±.00}                         & 0.5310\text{\scriptsize ±.00}                         & 0.1847\text{\scriptsize ±.02}                         \\
                                                                                              & Sim+FAGCN            & 0.7507\text{\scriptsize ±.01}                         & 0.9478\text{\scriptsize ±.00}                         & 0.7393\text{\scriptsize ±.03}                         & 0.8497\text{\scriptsize ±.00}                         & 0.7501\text{\scriptsize ±.00}                         & 0.8012\text{\scriptsize ±.01}                         & 0.8437\text{\scriptsize ±.02}                         & 0.9742\text{\scriptsize ±.00}                         & 0.8163\text{\scriptsize ±.02}                         & 0.2494\text{\scriptsize ±.02}                         & 0.5378\text{\scriptsize ±.00}                         & 0.2201\text{\scriptsize ±.00}                         & 0.2163\text{\scriptsize ±.00}                         & 0.5183\text{\scriptsize ±.00}                         & 0.1948\text{\scriptsize ±.00}                         \\ \midrule
\multicolumn{2}{c|}{IMP(\%)}                                                                                            & 5.96\%                                    & 1.90\%                                    & 5.08\%                                    & 6.13\%                                    & 2.46\%                                    & 10.87\%                                    & 6.07\%                                    & 2.37\%                                    & 5.63\%                                    & 0.82\%                                    & 0.59\%                                    & 1.28\%                                    & -1.21\%                                    & 1.04\%                                    & 2.92\%                                    \\
\cmidrule[1.2pt]{1-17}
\end{tabular}
}
\end{table*}

\begin{table*}[htbp]
\renewcommand\arraystretch{1.2}
\caption{Cross-domain transfer learning performance (mean±std Acc/AUC/F1) on homophilic datasets (C-way-1-shot) of GCOPE with different number of graph coordinators. `-1', `-3', and `-5' represent that we assign one, three, or five graph coordinators for each dataset in GCOPE.}
\label{appendix: Quantitative Analysis homo}
\resizebox{1.0\textwidth}{!}{
\begin{tabular}{c|ccc|ccc|ccc|ccc|ccc}
\cmidrule[1.2pt]{1-16}
\multirow{2}{*}{Methods} & \multicolumn{3}{c|}{Cora}                           & \multicolumn{3}{c|}{Citeseer}                       & \multicolumn{3}{c|}{Pubmed}                         & \multicolumn{3}{c|}{Computers}                      & \multicolumn{3}{c|}{Photo}                          \\
                         & Acc             & AUC             & F1              & Acc             & AUC             & F1              & Acc             & AUC             & F1              & Acc             & AUC             & F1              & Acc             & AUC             & F1              \\ \midrule
Supervised                & 0.3819\text{\scriptsize ±.03} & 0.6818\text{\scriptsize ±.04} & 0.3009\text{\scriptsize ±.09} & 0.5219\text{\scriptsize ±.08} & 0.8042\text{\scriptsize ±.03} & 0.4667\text{\scriptsize ±.08} & 0.4522\text{\scriptsize ±.02} & 0.5622\text{\scriptsize ±.04} & 0.4275\text{\scriptsize ±.07} & 0.4651\text{\scriptsize ±.04} & 0.7762\text{\scriptsize ±.02} & 0.3009\text{\scriptsize ±.07} & 0.5937\text{\scriptsize ±.05} & 0.8847\text{\scriptsize ±.00} & 0.5346\text{\scriptsize ±.03} \\ \midrule
IP                       & 0.3892\text{\scriptsize ±.05} & 0.7228\text{\scriptsize ±.03} & 0.3619\text{\scriptsize ±.05} & 0.4461\text{\scriptsize ±.02} & 0.7781\text{\scriptsize ±.01} & 0.4126\text{\scriptsize ±.02} & 0.4532\text{\scriptsize ±.02} & 0.5708\text{\scriptsize ±.03} & 0.4168\text{\scriptsize ±.04} & 0.4371\text{\scriptsize ±.06} & 0.7616\text{\scriptsize ±.01} & 0.3450\text{\scriptsize ±.02} & 0.6273\text{\scriptsize ±.01} & 0.8710\text{\scriptsize ±.01} & 0.5406\text{\scriptsize ±.03} \\ \midrule
GCOPE-1                  & 0.4618\text{\scriptsize ±.03} & 0.7597\text{\scriptsize ±.05} & 0.4388\text{\scriptsize ±.05} & 0.5631\text{\scriptsize ±.03} & 0.8258\text{\scriptsize ±.02} & 0.4953\text{\scriptsize ±.04} & 0.4591\text{\scriptsize ±.01} & 0.5512\text{\scriptsize ±.01} & 0.4203\text{\scriptsize ±.03} & 0.4465\text{\scriptsize ±.01} & 0.7747\text{\scriptsize ±.00} & 0.3432\text{\scriptsize ±.03} & 0.6329\text{\scriptsize ±.02} & 0.8850\text{\scriptsize ±.00} & 0.5935\text{\scriptsize ±.03} \\
GCOPE-3                  & 0.4272\text{\scriptsize ±.05} & 0.7509\text{\scriptsize ±.01} & 0.4176\text{\scriptsize ±.03} & 0.5518\text{\scriptsize ±.01} & 0.8438\text{\scriptsize ±.00} & 0.5074\text{\scriptsize ±.02} & 0.4777\text{\scriptsize ±.02} & 0.5689\text{\scriptsize ±.04} & 0.3794\text{\scriptsize ±.04} & 0.4314\text{\scriptsize ±.02} & 0.7335\text{\scriptsize ±.01} & 0.3546\text{\scriptsize ±.00} & 0.6491\text{\scriptsize ±.01} & 0.8934\text{\scriptsize ±.00} & 0.6109\text{\scriptsize ±.01} \\
GCOPE-5                  & 0.4499\text{\scriptsize ±.05} & 0.7673\text{\scriptsize ±.03} & 0.4414\text{\scriptsize ±.04} & 0.5280\text{\scriptsize ±.04} & 0.8183\text{\scriptsize ±.03} & 0.4658\text{\scriptsize ±.04} & 0.4686\text{\scriptsize ±.04} & 0.5737\text{\scriptsize ±.08} & 0.3663\text{\scriptsize ±.08} & 0.4650\text{\scriptsize ±.00} & 0.7656\text{\scriptsize ±.01} & 0.3767\text{\scriptsize ±.01} & 0.5538\text{\scriptsize ±.05} & 0.8760\text{\scriptsize ±.01} & 0.5621\text{\scriptsize ±.04} \\
\cmidrule[1.2pt]{1-16}
\end{tabular}
}
\end{table*}

\begin{table*}[htbp]
\renewcommand\arraystretch{1.2}
\caption{Cross-domain transfer learning performance (mean±std Acc/AUC/F1) on heterophilic datasets (C-way-1-shot) of GCOPE with different numbers of graph coordinators. `-1', `-3', and `-5' represents the we assign one, three, or five graph coordinators for each dataset in GCOPE.}
\label{appendix: Quantitative Analysis hetero}
\resizebox{1.0\textwidth}{!}{
\begin{tabular}{c|ccc|ccc|ccc|ccc|ccc}
\cmidrule[1.2pt]{1-16}
\multirow{2}{*}{Methods} & \multicolumn{3}{c|}{Wisconsin}                      & \multicolumn{3}{c|}{Texas}                          & \multicolumn{3}{c|}{Cornell}                        & \multicolumn{3}{c|}{Chameleon}                      & \multicolumn{3}{c|}{Squirrel}                       \\
                         & Acc             & AUC             & F1              & Acc             & AUC             & F1              & Acc             & AUC             & F1              & Acc             & AUC             & F1              & Acc             & AUC             & F1               \\ \midrule
Supervised                & 0.5222\text{\scriptsize ±.05} & 0.7905\text{\scriptsize ±.03} & 0.4725\text{\scriptsize ±.06} & 0.6900\text{\scriptsize ±.06} & 0.7185\text{\scriptsize ±.01} & 0.5334\text{\scriptsize ±.12} & 0.2938\text{\scriptsize ±.06} & 0.6573\text{\scriptsize ±.04} & 0.2872\text{\scriptsize ±.05} & 0.2575\text{\scriptsize ±.02} & 0.5515\text{\scriptsize ±.02} & 0.1941\text{\scriptsize ±.01} & 0.2181\text{\scriptsize ±.00} & 0.5202\text{\scriptsize ±.00} & 0.1875\text{\scriptsize ±.02} \\ \midrule
IP                       & 0.6063\text{\scriptsize ±.04} & 0.8356\text{\scriptsize ±.01} & 0.5555\text{\scriptsize ±.07} & 0.7425\text{\scriptsize ±.03} & 0.7034\text{\scriptsize ±.03} & 0.6141\text{\scriptsize ±.09} & 0.2588\text{\scriptsize ±.04} & 0.6262\text{\scriptsize ±.04} & 0.2442\text{\scriptsize ±.04} & 0.2443\text{\scriptsize ±.00} & 0.5530\text{\scriptsize ±.01} & 0.1875\text{\scriptsize ±.01} & 0.2223\text{\scriptsize ±.00} & 0.5307\text{\scriptsize ±.00} & 0.1740\text{\scriptsize ±.02} \\ \midrule
GCOPE-1                  & 0.6579\text{\scriptsize ±.03} & 0.8531\text{\scriptsize ±.01} & 0.5649\text{\scriptsize ±.00} & 0.7125\text{\scriptsize ±.02} & 0.6693\text{\scriptsize ±.02} & 0.6300\text{\scriptsize ±.03} & 0.4013\text{\scriptsize ±.05} & 0.6897\text{\scriptsize ±.01} & 0.3160\text{\scriptsize ±.02} & 0.2886\text{\scriptsize ±.00} & 0.5898\text{\scriptsize ±.00} & 0.2320\text{\scriptsize ±.00} & 0.2257\text{\scriptsize ±.00} & 0.5257\text{\scriptsize ±.00} & 0.1885\text{\scriptsize ±.01} \\
GCOPE-3                  & 0.6217\text{\scriptsize ±.00} & 0.8267\text{\scriptsize ±.00} & 0.5397\text{\scriptsize ±.01} & 0.7675\text{\scriptsize ±.04} & 0.7005\text{\scriptsize ±.03} & 0.5834\text{\scriptsize ±.05} & 0.5675\text{\scriptsize ±.03} & 0.7334\text{\scriptsize ±.01} & 0.4506\text{\scriptsize ±.02} & 0.2895\text{\scriptsize ±.00} & 0.5785\text{\scriptsize ±.00} & 0.2205\text{\scriptsize ±.00} & 0.2199\text{\scriptsize ±.01} & 0.5274\text{\scriptsize ±.01} & 0.1815\text{\scriptsize ±.02} \\
GCOPE-5                  & 0.6353\text{\scriptsize ±.06} & 0.8327\text{\scriptsize ±.03} & 0.5692\text{\scriptsize ±.05} & 0.7262\text{\scriptsize ±.04} & 0.6913\text{\scriptsize ±.02} & 0.5468\text{\scriptsize ±.04} & 0.6550\text{\scriptsize ±.02} & 0.8350\text{\scriptsize ±.00} & 0.5047\text{\scriptsize ±.02} & 0.2810\text{\scriptsize ±.00} & 0.5654\text{\scriptsize ±.00} & 0.2113\text{\scriptsize ±.00} & 0.2129\text{\scriptsize ±.00} & 0.5183\text{\scriptsize ±.00} & 0.2029\text{\scriptsize ±.00} \\
\cmidrule[1.2pt]{1-16}
\end{tabular}
}
\end{table*}

\begin{table*}[htbp]
\caption{Cross-domain transfer learning performance (mean±std Acc/AUC/F1) on two homophilic and two heterophilic datasets (C-way-1-shot) of GCOPE with full or dynamical inter-coordinator edges. '/f' means that the inter-coordinator edges in GCOPE are fully connected to each other. '/d' represents that inter-dataset edges in GCOPE are dynamically connected by computing the similarity between them.}
\label{appendix: dynamical inter-dataset edges}
\renewcommand\arraystretch{1.2}
\resizebox{1.0\textwidth}{!}{
\begin{tabular}{c|ccc|ccc|ccc|ccc}
\cmidrule[1.2pt]{1-13}
\multirow{2}{*}{Methods} & \multicolumn{3}{c|}{Cora}                           & \multicolumn{3}{c|}{Citeseer}                       & \multicolumn{3}{c|}{Wisconsin}                      & \multicolumn{3}{c}{Texas}                          \\
                         & Acc             & AUC             & F1              & Acc             & AUC             & F1              & Acc             & AUC             & F1              & Acc             & AUC             & F1              \\ \midrule
Supervised               & 0.3819\text{\scriptsize ±.03} & 0.6818\text{\scriptsize ±.04} & 0.3009\text{\scriptsize ±.09} & 0.5219\text{\scriptsize ±.08} & 0.8042\text{\scriptsize ±.03} & 0.4667\text{\scriptsize ±.08} & 0.5222\text{\scriptsize ±.05} & 0.7905\text{\scriptsize ±.03} & 0.4725\text{\scriptsize ±.06} & 0.6900\text{\scriptsize ±.06} & 0.7185\text{\scriptsize ±.01} & 0.5334\text{\scriptsize ±.12} \\ \midrule
GCOPE/f                  & 0.4618\text{\scriptsize ±.03} & 0.7597\text{\scriptsize ±.05} & 0.4388\text{\scriptsize ±.05} & 0.5631\text{\scriptsize ±.03} & 0.8258\text{\scriptsize ±.02} & 0.4953\text{\scriptsize ±.04} & 0.6579\text{\scriptsize ±.03} & 0.8531\text{\scriptsize ±.01} & 0.5649\text{\scriptsize ±.00} & 0.7125\text{\scriptsize ±.02} & 0.6693\text{\scriptsize ±.02} & 0.6300\text{\scriptsize ±.03} \\ \midrule
GCOPE/d                  & 0.4027\text{\scriptsize ±.00} & 0.7228\text{\scriptsize ±.01} & 0.4206\text{\scriptsize ±.00} & 0.4874\text{\scriptsize ±.04} & 0.7764\text{\scriptsize ±.03} & 0.4342\text{\scriptsize ±.04} & 0.6588\text{\scriptsize ±.01} & 0.8545\text{\scriptsize ±.00} & 0.6046\text{\scriptsize ±.02} & 0.7400\text{\scriptsize ±.03} & 0.6851\text{\scriptsize ±.02} & 0.5813\text{\scriptsize ±.05} \\
\cmidrule[1.2pt]{1-13}
\end{tabular}
}
\end{table*}

\section{Appendix}
\label{appendix: exps}

\subsection{Codes and Datasets}
\textbf{Code available at \url{https://github.com/cshhzhao/GCOPE}}. A more detailed information on related datasets is as follows:  
\begin{itemize}
    \item \textbf{Citation network:} 
    Cora and Citeseer datasets \cite{sen2008homoCoraCiteseer} consist of a diverse collection of computer science publications, where each node is characterized by a bag-of-words representation of papers and a categorical label indicating the paper topic. Pubmed dataset \cite{namata2012homoPubmed} comprises articles related to diabetes from the PubMed database. Each node in this dataset is represented by an attribute vector containing TF/IDF-weighted word frequencies, accompanied by a label specifying the particular type of diabetes discussed in the publication.
    \item \textbf{Amazon network:} 
    Computers and Photos datasets \cite{mcauley2015homoComputerPhoto,shchur2018homoComputersPhoto} are two networks illustrating co-purchase relationships sourced from Amazon. In these networks, each node represents a product, and an edge indicates frequent co-purchases between two products. Additionally, each node is associated with a bag-of-words representation of product reviews and is labeled with its respective category.
    \item  \textbf{WebKB:} 
    Cornell, Texas, and Wisconsin are three subdatasets derived from the WebKB dataset~\cite{pei2019heterophilicDatasets}, compiled from multiple universities' web pages. Each node within these datasets represents a web page, with edges denoting hyperlinks between pages. The node features are represented as bag-of-words representations of the web pages. Additionally, the web pages are manually categorized into five distinct labels: student, project, course, staff, and faculty.
    \item \textbf{Wikipedia network:} 
    Chameleon and Squirrel datasets \cite{pei2019heterophilicDatasets} consist of two page-page networks extracted from Wikipedia, focusing on specific topics. In these networks, nodes represent individual web pages, while edges signify links between pages. Node attributes are defined as sets of informative nouns extracted from the pages. Moreover, each node is labeled based on the average monthly traffic received by the respective web page.
\end{itemize}

\vspace{-1.25em}
\subsection{Parameter Analysis}
\vspace{-0.25em}
We compare parameter numbers of two baselines and our proposed GCOPE in Table~\ref{tab: parameter analysis}. For downstream tasks, we also compare parameter numbers of finetuning and prompting in Table~\ref{tab: parameter analysis}.

\newenvironment{tablehere} 
    {\def\@captype{table}} 
    {} 
\makeatother
\makeatletter
\def\@captype{table}
\makeatother

\begin{center}
\setlength{\tabcolsep}{2.0mm}
\centering
\vspace{-1.25em}
\caption{
Parameter analysis various methods, where the backbone is FAGCN. IP and GCOPE are pretrained by GraphCL. Citeseer is the downstream dataset. We use ``-'' to represent the case without corresponding phases.
}
\label{tab: parameter analysis}
\resizebox{0.3\textwidth}{!}{
\begin{tabular}{c|c|c|ccc}
\cmidrule[1.2pt]{1-4}
Methods          & Supervised & IP    & GCOPE \\ \midrule
Pretraining      & -          & 62,976 & 62,976 \\ \midrule
+ Coordinators   & -          & -     & 900   \\ \midrule
+ Reconstruction & -          & -     & 29,412 \\ 
\midrule
\midrule
Finetuning       & 47,750      & 47,750 & 47,750 \\ \midrule
Prompting        & -          & -     & 19,398 \\ 
\cmidrule[1.2pt]{1-4}
\end{tabular}
}
\label{table:dataset_appendix_1}
\end{center}

\vspace{-1.25em}
\subsection{Cross-domain Transfer Performance with other C-way-K-shot settings}
\vspace{-0.25em}
We use Table~\ref{tab:rq1 homo 3 shot}, Table~\ref{tab:rq1 hetero 3 shot}, Table~\ref{tab:rq1 homo 5 shot}, and Table~\ref{tab:rq1 hetero 5 shot} to show the cross-domain transfer learning performance with 3/5-shot learning scenarios respectively. From the experimental results, we observe that our proposed GCOPE maintains its superiority over other baseline methods, even though the improvements modestly decrease from 1-shot to 5-shot scenarios.

\vspace{-1.25em}
\subsection{Quantitative Analysis of Graph Coordinators}
\vspace{-0.25em}
We conduct additional experiments to investigate the quantitative analysis of graph coordinators, shown by Table~\ref{appendix: Quantitative Analysis homo} and Table~\ref{appendix: Quantitative Analysis hetero}. Experimental results show that, despite slight fluctuations in performance across different datasets, GCOPE with varying numbers of graph coordinators generally outperforms both the Supervised and IP methods.

\vspace{-1.em}
\subsection{Dynamical Study of inter-coordinator edges}
We additionally study the impact of dynamical inter-coordinator edges on the performance of GCOPE, and report the transfer learning results in Table~\ref{appendix: dynamical inter-dataset edges}. According to the experimental results, we observe that GCOPE/d outperforms GCOPE/f on the Wisconsin and Texas, performs better than the supervised method on the Cora but is inferior to GCOPE/f, and exhibits negative transfer on Citeseer. In comparison, GCOPE/f demonstrates positive transfer across all datasets.


\end{document}